\documentclass{article}
\usepackage[dvipdfmx]{graphicx}
\usepackage{arxiv}
\usepackage[utf8]{inputenc} 
\usepackage[T1]{fontenc}    
\usepackage{hyperref}       
\usepackage{url}            
\usepackage{booktabs}       
\usepackage{amsfonts}       
\usepackage{nicefrac}       
\usepackage{microtype}      
\usepackage{bm}
\usepackage{lipsum}
\usepackage{amsmath}
\usepackage{algorithm}
\usepackage{algorithmic}
\usepackage{multirow}
\usepackage{comment}
\newcommand{\detector}{ONLAD}
\newcommand{\core}{{\detector} Core}
\newcommand{\noff}{{\detector}-NF}
\newcommand{\nn}{NN-AE}
\newcommand{\dnn}{DNN-AE}
\newcommand{\fpelm}{FPELM-AE}
\newcommand{\nncpu}{{\nn}-CPU}
\newcommand{\nngpu}{{\nn}-GPU}
\newcommand{\dnncpu}{{\dnn}-CPU}
\newcommand{\dnngpu}{{\dnn}-GPU}
\newcommand{\fpelmcpu}{{\fpelm}-CPU}
\newcommand{\fpelmgpu}{{\fpelm}-GPU}

\title{A Neural Network-Based On-device Learning Anomaly Detector for Edge Devices}
\author{
  Mineto Tsukada\\
  Keio University\\
  3-14-1 Hiyoshi, Kohoku-ku, Yokohama, Japan\\
  \texttt{tsukada@arc.ics.keio.ac.jp}\\
  \And
  Masaaki Kondo \\
  The University of Tokyo\\
  7-3-1 Hongo, Bunkyo-ku, Tokyo, Japan\\
  \texttt{kondo@hal.ipc.i.u-tokyo.ac.jp} \\
  \And
  Hiroki Matsutani \\
  Keio University\\
  3-14-1 Hiyoshi, Kohoku-ku, Yokohama, Japan\\
  \texttt{matutani@arc.ics.keio.ac.jp} \\
}

\begin{document}
\maketitle

\begin{abstract}
  Semi-supervised anomaly detection is an approach to identify anomalies by learning the distribution of normal data.
  Backpropagation neural networks (i.e., BP-NNs) based approaches have recently drawn attention
  because of their good generalization capability.
  In a typical situation, BP-NN-based models are iteratively optimized in server machines with input data gathered from edge devices.
  However, (1) the iterative optimization often requires significant efforts to follow changes in the distribution of normal data (i.e., concept drift), and
  (2) data transfers between edge and server impose additional latency and energy consumption.
  To address these issues, we propose {\detector} and its IP core, named {\core}.
  {\detector} is highly optimized to perform fast sequential learning
  to follow concept drift in less than one millisecond.
  {\core} realizes on-device learning for edge devices at low power consumption,
  which realizes standalone execution where data transfers between edge and server are not required.
  Experiments show that {\detector} has favorable anomaly detection capability in an environment that simulates concept drift.
  Evaluations of {\core} confirm that the training latency is 1.95x$\sim$6.58x faster than the other software implementations.
  Also, the runtime power consumption of {\core} implemented on PYNQ-Z1 board, a small FPGA/CPU SoC platform,
  is 5.0x$\sim$25.4x lower than them.
\end{abstract}

\section{Introduction}\label{sec:intro}
Anomaly detection is an approach to identify rare data instances (i.e., anomalies)
that have different patterns or come from different distributions from that of the majority (i.e., the normal class) \cite{anomaly_detection_survey}.
There are mainly three approaches in anomaly detection: (1) supervised anomaly detection, (2) semi-supervised anomaly detection, and (3) unsupervised anomaly detection.

(1) A typical strategy of supervised anomaly detection is
to build a binary-classification model for the normal class vs. the anomaly class \cite{anomaly_detection_survey}.
It requires labeled normal and anomaly data to train a model,
however, anomaly instances are basically much rarer than normal ones,
which imposes the class-imbalanced problem \cite{class_imbalance}.
Several works have addressed this issue by undersampling the majority data
or oversampling the minority data \cite{lewis_resample}\cite{andrew_multiple_resample},
or assigning more costs on misclassified data to make the classifier concentrate the
minority classes \cite{pazzani_cost_sensitive}.

(2) Semi-supervised anomaly detection, one of the main topics of this paper, assumes that
all the training data belong to the normal class \cite{anomaly_detection_survey}.
A typical strategy of semi-supervised anomaly detection is to learn the distribution of normal data
and then to identify data samples distant from the distribution as anomalies.
Semi-supervised approaches do not require anomalies to train a model,
which makes them applicable to a wide range of real-world tasks.
Various approaches have been proposed, such as
nearest-neighbor based techniques \cite{lof}\cite{knn_anoma_liao},
clustering approaches \cite{cluster_anoma_leung}\cite{cluster_anoma_munz},
and one-class classification approaches \cite{one_class_svm_anoma_wang}\cite{one_class_svm_anoma_li}.

(3) Unsupervised anomaly detection does not require labeled training data \cite{anomaly_detection_survey},
thus its constraint is the least restrictive.
Many semi-supervised methods can be used in an unsupervised manner
by using unlabeled data to train a model because most unlabeled data belong to the normal class.
Sometimes, unsupervised anomaly detection and semi-supervised anomaly detection
are not distinguished explicitly.

In this paper, we focus on semi-supervised anomaly detection.
Recently, neural network-based approaches \cite{one_class_deep_svm_anoma}\cite{ae_anoma}\cite{gan_anoma} have been drawing attention because
in many cases they achieve relatively higher generalization performance than the traditional approaches for a wide range of real-world data such as images, natural languages, and audio data.
Although there are some variants of neural networks,
backpropagation neural networks (i.e., BP-NNs) are currently widely used.
\begin{figure}[t]
    \centering
    \begin{minipage}{0.475\hsize}
        \centering
        \includegraphics[height=37.5mm]{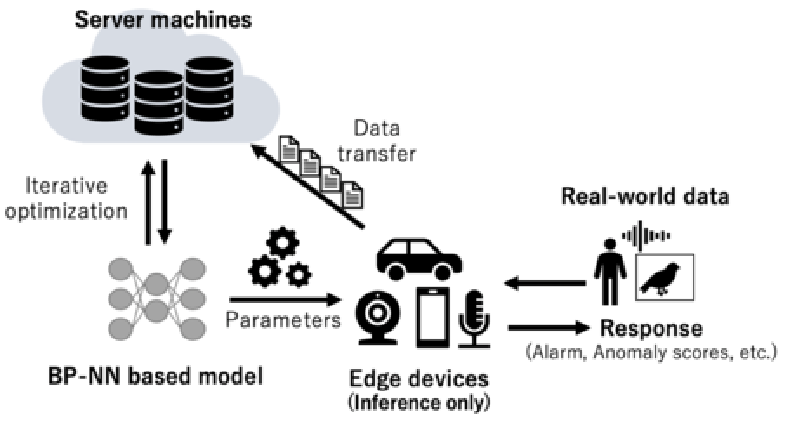}
        \caption{Typical Application of BP-NN-based Semi-supervised Anomaly Detection Models}
        \label{fig:offline}
    \end{minipage}
    \begin{minipage}{0.475\hsize}
        \centering
        \includegraphics[height=37.5mm]{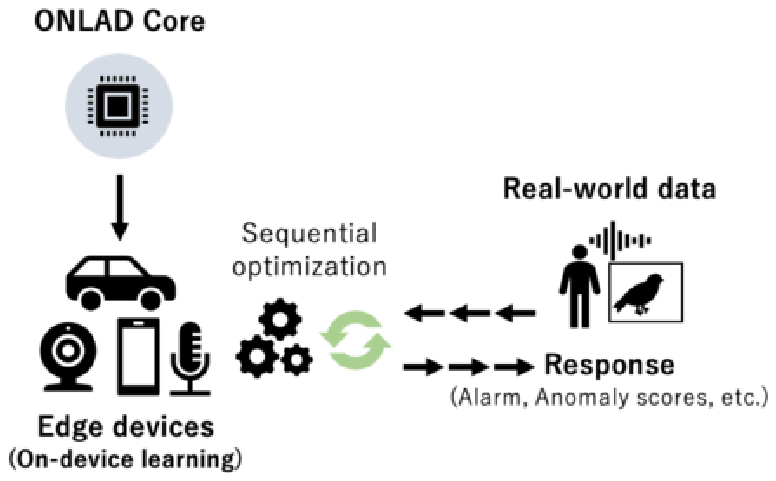}
        \caption{On-device Sequential Learning Approach}
        \label{fig:ondevice}
    \end{minipage}
\end{figure}

Figure \ref{fig:offline} illustrates a typical application of BP-NN-based semi-supervised anomaly detection models.
The system shown in the figure is designed for edge devices that implement their own models to detect anomalies of incoming real-world data.
In this system, the edge devices are supposed to perform only inference computations (e.g., calculating anomaly scores),
and training computations are offloaded to server machines.
The models are iteratively trained in the server machines with a large amount of input data gathered from the edge devices.
Once the training loop completes, parameters of the edge devices are updated with the optimized ones.
However, there are two issues with this approach:
(1) BP-NNs' iterative optimization approach
often takes a considerable computation time,
which makes it difficult to follow time-series changes
in the distribution of normal data (i.e., concept drift).
(2) Data transfers to the server machines may impose
several problems on the edge devices such as
additional latency and energy consumption for communication.

(1) As mentioned before, learning the distribution of normal data
is a key feature of semi-supervised anomaly detection approaches.
However, the distribution may change over time.
This phenomenon is referred to as concept drift.
Concept drift is a serious problem when there are frequent changes
in the surrounding environment of data \cite{cd_environment}
or behavioral state changes in data sources \cite{cd_user_model}.
A semi-supervised anomaly detection model
should learn new normal data to follow the changes,
however, BP-NNs' iterative optimization approach often introduces a considerable delay,
which widens a gap between the latest true distribution of normal data
and the one learned by the model \cite{trusted_ai_ibm}.
This gap makes identifying anomalies more difficult gradually.

(2) Usually, edge devices that implement machine learning models are specialized only for prediction computations because
the backpropagation method often requires a large amount of computational power.
This is why training computations of BP-NNs are typically offloaded to server machines with high computational power.
In this case, data transfers to the server machines are inevitable,
which imposes additional energy consumption for communication and potential risk of data breaches on the edge devices.

One practical solution to these two issues is the on-device sequential learning approach illustrated in Figure \ref{fig:ondevice}.
In this approach, incoming input data are sequentially learned on edge devices themselves.
This approach allows the edge devices to sequentially follow changes in the distribution of normal data and
makes possible standalone execution where no data transfers are required.
However, it poses challenges in regard to how to construct such a sequential learning algorithm
and how to implement it on edge devices with limited resources.

To deal with the underlying challenges, we propose an ON-device sequential Learning semi-supervised Anomaly Detector called {\detector}
and its IP core, named {\core}
\footnote{
    This work is an extended version of our prior work \cite{os_elm_fpga_tsukada}.
}.
The algorithm of {\detector} is designed to perform fast sequential learning to follow concept drift in less than one millisecond.
{\core} realizes on-device learning for resource-limited edge devices at low power consumption.

In this paper, we make the following contributions:
\begin{enumerate}
    \item {\detector} leverages OS-ELM \cite{os_elm}, a lightweight neural network that can perform fast sequential learning, as a core component.
    In Section \ref{ssec:detector_anal}, we theoretically analyze the training algorithm of OS-ELM and
    demonstrate that the computational cost significantly reduces without degrading the training results when the batch size equals 1.
    \item In Section \ref{ssec:detector_forget}, we propose a computationally lightweight forgetting mechanism for OS-ELM
    based on FP-ELM, a state-of-the-art OS-ELM variant with a dynamic forgetting mechanism.
    Since a key feature of semi-supervised anomaly detection is to learn the distribution of normal data,
    OS-ELM should be able to forget past learned normal data when the distribution changes.
    The proposed method provides such a function for OS-ELM with a tiny additional computational cost.
    \item In Section \ref{ssec:detector_algo}, we propose {\detector}, a new sequential learning semi-supervised anomaly detector that combines OS-ELM and an autoencoder \cite{autoencoder},
    a neural network-based dimensionality reduction model.
    This combination, together with the other proposed techniques to reduce the computational cost,
    realizes fast sequential learning semi-supervised anomaly detection.
    Experiments using several public datasets in Section \ref{sec:eval_anomaly} show that
    {\detector} has comparable generalization capability to that
    of BP-NN based models in the context of anomaly detection.
    They also confirm that {\detector} outperforms BP-NN based models
    in terms of anomaly detection capability especially in an environment that simulates concept drift.
    \item In Section \ref{sec:imple}, we describe the design and implementation of {\core}.
    Evaluations of {\core} in Section \ref{sec:eval_hard} show that {\core} can perform training and prediction computations approximately in less than one millisecond.
    In comparison with software counterparts,
    the training latency of {\core} is faster by 1.95x$\sim$6.58x,
    while the prediction latency is faster by 2.29x$\sim$4.73x on average.
    They also confirm that the proposed forgetting mechanism is faster
    than the baseline algorithm, FP-ELM, by 3.21x on average.
    In addition, our evaluations show that {\core} can be implemented on PYNQ-Z1 board, a small FPGA/CPU SoC platform,
    in practical model sizes.
    It is also demonstrated that the runtime power consumption
    of PYNQ-Z1 board that implements {\core} is 5.0x$\sim$25.4x lower than the other software counterparts
    when training computations are continuously executed.
\end{enumerate}

The rest of this paper is organized as follows: Section \ref{sec:prelim} provides a brief review of the basic technologies behind {\detector}.
We propose {\detector} in Section \ref{sec:design}.
Section \ref{sec:imple} describes the design and implementation of {\core}.
{\detector} is evaluated in terms of anomaly detection capability in Section \ref{sec:eval_anomaly}.
{\core} is also evaluated in terms of latency, FPGA resource utilization, and power consumption in Section \ref{sec:eval_hard}.
Related works are described in Section \ref{sec:related}.
Section \ref{sec:conc} concludes this paper.

\section{Preliminaries}\label{sec:prelim}
This section provides a brief introduction of the base technologies behind {\detector}:
(1) ELM (Extreme Learning Machine), (2) OS-ELM (Online Sequential Extreme Learning Machine), and (3) autoencoders.

\subsection{ELM}\label{ssec:elm}
\begin{figure}[t]
    \centering
    \includegraphics[width=80.0mm]{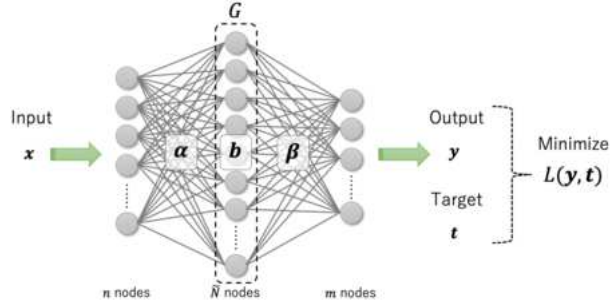}
    \caption{Extreme Learning Machine}
    \label{fig:elm}
\end{figure}

ELM \cite{elm} illustrated in Figure \ref{fig:elm} is a kind of single hidden layer feedforward network (i.e., SLFN)
that consists of an input layer, a hidden layer, and an output layer.
Suppose an $n$-dimensional input chunk $\bm{x} \in \bm{R}^{k \times n}$ of batch size $= k$ is given;
an $m$-dimensional output chunk $\bm{y} \in \bm{R}^{k \times m}$ is computed as follows.
\begin{equation}\label{eq:elm_predict}
    \bm{y} = G(\bm{x} \cdot \bm{\alpha} + \bm{b})\bm{\beta},
\end{equation}
where $\bm{\alpha} \in \bm{R}^{n \times \tilde{N}}$ denotes an input weight connecting
the input layer and the hidden layer, and $\bm{\beta} \in \bm{R}^{\tilde{N} \times m}$
an output weight connecting the hidden layer and the output layer.
$\bm{b} \in \bm{R}^{\tilde{N}}$ denotes a bias vector of the hidden layer,
and $G$ an activation function applied to the hidden layer output.

If an SLFN can approximate an $m$-dimensional target chunk $\bm{t} \in \bm{R}^{k \times m}$
with zero error, it implies that there exists $\bm{\beta}$ which satisfies the following equation.
\begin{equation}\label{eq:elm_zero_error}
    G(\bm{x} \cdot \bm{\alpha} + \bm{b})\bm{\beta} = \bm{t}
\end{equation}
Let $\bm{H} \in \bm{R}^{k \times \tilde{N}}$ be the hidden layer output $G(\bm{x} \cdot \bm{\alpha} + \bm{b})$;
then the optimal output weight $\hat{\bm{\beta}}$ is computed as follows.
\begin{equation}\label{eq:elm_train}
    \hat{\bm{\beta}} = \bm{H}^{\dagger}\bm{t},
\end{equation}
where $\bm{H}^{\dagger}$ is the pseudo inverse of $\bm{H}$.
$\bm{H}^{\dagger}$ can be calculated with matrix decomposition algorithms such as SVD (Singular Value Decomposition) \cite{svd}.
In particular, if $\bm{H}^T\bm{H}$ or $\bm{H}\bm{H}^T$ is non-singular, $\bm{H}^{\dagger}$ can be calculated in an efficient way with
$\bm{H}^{\dagger} = (\bm{H}^T\bm{H})^{-1}\bm{H}^T$ or $\bm{H}^{\dagger} = \bm{H}^T(\bm{H}\bm{H}^T)^{-1}$.

The whole training process is completed simply by replacing $\bm{\beta}$ with $\hat{\bm{\beta}}$.
$\bm{\alpha}$ and $\bm{b}$ do not change once they have been initialized with random values;
the conversion from $\bm{x}$ to $\bm{H}$ is random projection.

ELM does not use iterative optimization that BP-NNs use,
but rather one-shot optimization, which makes the whole training process faster. 
ELM can compute the optimal output weight faster than BP-NNs \cite{elm}.
It is categorized as a batch learning algorithm, wherein
all the training data are assumed to be available in advance.
In other words, ELM must be retrained with the whole dataset, including the past training data,
in order to learn new instances.

\subsection{OS-ELM}\label{ssec:oselm}
OS-ELM \cite{os_elm} is an ELM variant that can perform sequential learning instead of batch learning.
Suppose the $i$th training chunk $\{\bm{x}_i \in \bm{R}^{k_i \times n}, \bm{t}_i \in \bm{R}^{k_i \times m}\}$
of batch size $= k_i$ is given; we need to find $\bm{\beta}$ that minimizes the following error.
\begin{equation}\label{eq:os_elm_error}
    \begin{Vmatrix}
        \begin{bmatrix}\bm{H}_0 \\ \vdots \\ \bm{H}_i\end{bmatrix} \bm{\beta}_i - \begin{bmatrix} \bm{t}_0 \\ \vdots \\ \bm{t}_i \end{bmatrix}
    \end{Vmatrix},
\end{equation}
where $\bm{H}_i$ is defined as $\bm{H}_i \equiv G(\bm{x}_i \cdot \bm{\alpha} + \bm{b})$.
The optimal output weight is sequentially computed as follows.
\begin{equation}\label{eq:os_elm_train}
    \begin{split}
        \bm{P}_i &= \bm{P}_{i-1} - \bm{P}_{i-1}\bm{H}_i^T(\bm{I} + \bm{H}_i\bm{P}_{i-1}\bm{H}_i^T)^{-1}\bm{H}_i\bm{P}_{i-1}\\
        \bm{\beta}_i &= \bm{\beta}_{i-1} + \bm{P}_i\bm{H}_i^T(\bm{t}_i - \bm{H}_i\bm{\beta}_{i-1})
    \end{split}
\end{equation}
$\bm{P}_0$ and $\bm{\beta}_0$ are computed as follows.
\begin{equation}\label{eq:os_elm_init}
    \begin{split}
        \bm{P}_0 &= (\bm{H}_0^T\bm{H}_0)^{-1}\\
        \bm{\beta}_0 &= \bm{P}_0\bm{H}_0^T\bm{t}_0
    \end{split}
\end{equation}
The number of initial training samples $k_0$ should be greater than
that of hidden nodes $\tilde{\bm{N}}$ to make $\bm{H}_0^T\bm{H}_0$ nonsingular.

As shown in Equation \ref{eq:os_elm_train}, OS-ELM sequentially finds the optimal output weight for the new training chunk
without memory or retraining using past training data, unlike ELM.
OS-ELM can also find the optimal solution faster than BP-NNs \cite{os_elm}.

\subsection{Autoencoders}\label{ssec:ae}
\begin{figure}[t]
    \centering
    \includegraphics[width=80.0mm]{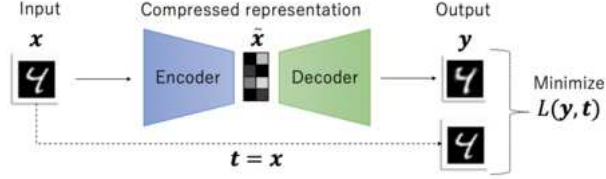}
    \caption{Autoencoder}
    \label{fig:autoencoder}
\end{figure}
An autoencoder \cite{autoencoder} illustrated in Figure \ref{fig:autoencoder} is a neural network-based unsupervised learning model
for finding a well-characterized dimensionality reduced form $\bm{\tilde{x}} \in \bm{R}^{k \times \tilde{n}}$ of an input chunk $\bm{x} \in \bm{R}^{k \times n}$ ($\tilde{n} < n$).
Generally, the output of an intermediate layer is regarded as $\bm{\tilde{x}}$.
ELM and OS-ELM have only one intermediate layer;
therefore, the hidden layer output $\bm{H}$ is regarded as $\bm{\tilde{x}}$.
Basically, the number of hidden nodes $\tilde{n}$ is constrained to be less than that
of input nodes $n$. Such autoencoders are specially referred to as undercomplete autoencoders.
However, sometimes they take the opposite setting (i.e., $n < \tilde{n}$) where they are referred to as overcomplete autoencoders.
Although overcomplete autoencoders cannot perform dimensionality reduction,
they can obtain well-characterized representations for classification problems by applying regularization conditions or noise \cite{denoae}
to their loss functions.

In the training process, input data are also used as targets (i.e., $\bm{t} = \bm{x}$);
therefore, an autoencoder is trained to correctly reconstruct input data as output data.
It is empirically known that $\bm{\tilde{x}}$ tends to become well-characterized when
the error between input data and reconstructed output data converges \cite{autoencoder}.
Labeled data are not required during the whole training process;
this is why an autoencoder is categorized as an unsupervised learning model.

Autoencoders have been attracting attention in the field of semi-supervised anomaly detection \cite{ae_anoma}\cite{vae_anoma}, too.
In this context, an autoencoder is trained only with normal data;
therefore, its output tends to have a relatively large reconstruction error in the case of an anomaly.
Thus, anomalies can be detected by setting a threshold for the errors.
This approach is categorized as a semi-supervised anomaly detection method since
only normal data are used as training data.

PCA (Principal Component Analysis), another non-statistical dimensionality reduction algorithm, is often compared with autoencoders.
Sakurada \textit{et al.} showed that autoencoder-based models can detect subtle
anomalies that PCA fails to pick up \cite{ae_anoma}.
Moreover, autoencoders can perform nonlinear transformations without costly computations that
kernel PCA \cite{kernel_pca} requires.
\section{{\detector}}\label{sec:design}
As mentioned in the introduction, {\detector} leverages OS-ELM as its core component.
In this section, we provide a theoretical analysis of OS-ELM and
demonstrate that the computational cost of the training algorithm significantly reduces when batch size = 1 without any deterioration of the training results.
Then, we propose a computationally lightweight forgetting mechanism to deal with concept drift.
Finally, we formulate the algorithm of {\detector}.

\subsection{Analysis of OS-ELM}\label{ssec:detector_anal}
The training algorithm of OS-ELM (i.e., Equation \ref{eq:os_elm_train}) mainly consists of (1) matrix products and (2) matrix inversions.
Suppose the computational iterations of a matrix product
$\bm{A} \in \bm{R^{p \times q}} \cdot \bm{B} \in \bm{R^{q \times r}}$ are $pqr$
and those of a matrix inversion $\bm{C}^{-1} \in \bm{R^{r \times r}}$ are $r^3$;
the total computational iterations of these two operations in Equation \ref{eq:os_elm_train} are calculated as follows.
\begin{equation}\label{eq:anal_prod_and_inv}
    \begin{split}
        I_{prod} &= 4k\tilde{N}^2 + k(2k + 2m + n)\tilde{N} \nonumber \\
        I_{inv} &= k^3,
    \end{split}
\end{equation}
where $I_{prod}$ denotes the total computational iterations of the matrix products,
while $I_{inv}$ denotes those of the matrix inversions.
$n$, $\tilde{N}$, and $m$ are the numbers of input, hidden, and output nodes of OS-ELM, respectively.
$k$ denotes the batch size.
For instance, the computational iterations of
$\bm{H}_i\bm{P}_{i-1}\bm{H}_i^T$ are calculated by dividing the computing process into two steps: 
(1) $\bm{H}_i \in \bm{R}^{k \times \tilde{N}} \cdot \bm{P}_{i-1} \in \bm{R}^{\tilde{N} \times \tilde{N}}$
and (2) $\bm{H}_i\bm{P}_{i-1} \in \bm{R}^{k \times \tilde{N}} \cdot \bm{H}_i^T \in \bm{R}^{\tilde{N} \times k}$.
In this case, these computational iterations are calculated as $k\tilde{N}^2$ and $k^2\tilde{N}$, respectively.

Let $I_k$ be the total computational iterations of
matrix products and matrix inversions in Equation \ref{eq:os_elm_train} when
batch size = $k$. Accordingly, the following equations can be derived.
\begin{equation}\label{eq:anal}
    \begin{split}
        I_k &= I_{prod} + I_{inv} \nonumber \\
        &= 4k\tilde{N}^2 + k(2k + 2m + n)\tilde{N} + k^3 \nonumber \\
        &= k(4\tilde{N}^2 + (2k + 2m + n)\tilde{N} + k^2) \nonumber \\
        &\ge k(4\tilde{N}^2 + (2 + 2m + n)\tilde{N} + 1) = kI_1
    \end{split}
\end{equation}
Finally, $I_k \geq kI_1$ is obtained.
This inequality shows that the training algorithm becomes computationally more efficient when batch size = 1, rather than when batch size = $k$ $(> 1)$.
Please note that this insight does not always make sense especially for software implementations because this computational model does not take into account the software-specific overheads such as memory allocation and function calls.
However, bare-metal implementations, including {\core}, receive benefits from this insight since they are free from such overheads.
Moreover, when $k$ = 1, the computational cost of the matrix inversion $(\bm{I} + \bm{H}_i\bm{P}_{i-1}\bm{H}_i^T)^{-1}$ in Equation \ref{eq:os_elm_train}
is significantly reduced, as the size of the target matrix $\bm{I} + \bm{H}_i\bm{P}_{i-1}\bm{H}_i^T$ is $k \times k$.
In this case, the following training algorithm is derived from Equation \ref{eq:os_elm_train}.
\begin{eqnarray}\label{eq:os_elm_train_bsize_1}
    \begin{split}
        \bm{P}_i &= \bm{P}_{i-1} - \frac{\bm{P}_{i-1}\bm{h}_i^T\bm{h}_i\bm{P}_{i-1}}{1+\bm{h}_i\bm{P}_{i-1}\bm{h}_i^T} \\
        \bm{\beta}_i &= \bm{\beta}_{i-1} + \bm{P}_i\bm{h}_i^T(\bm{t}_i - \bm{h}_i\bm{\beta}_{i-1}),
    \end{split}
\end{eqnarray}
where $\bm{h} \in \bm{R}^{\tilde{N}}$ denotes the special case of $\bm{H} \in \bm{R}^{k \times \tilde{N}}$ when $k$ = 1.
Thanks to the above trick, OS-ELM can perform training without any costly matrix inversions,
which helps to reduce not only the computational cost but also the hardware resources needed for {\core}.
It also makes it easier to parallelize the training algorithm, because
there are no matrix inversions with a low degree of parallelism in Equation \ref{eq:os_elm_train_bsize_1}.
Furthermore, the training results of OS-ELM are not affected even when batch size = 1,
because OS-ELM gives the same output weight when training is performed $N$ times with batch size = $k$ or $Nk$ times with batch size = 1.
This is a notable difference from BP-NNs; their training results get better or worse depending on the batch size.
On the basis of the above discussion, the batch size of OS-ELM used in {\detector} is always set to 1.

\subsection{Lightweight Forgetting Mechanism for OS-ELM}\label{ssec:detector_forget}
In certain real environments, the distribution of normal data may change as time goes by.
In this case, {\detector} should have a function to adaptively forget past learned normal data with a tiny additional computational cost.
To deal with this challenge, we propose a computationally lightweight forgetting mechanism based on
FP-ELM (Forgetting Parameters Extreme Learning Machine) \cite{fp_elm},
a state-of-the-art OS-ELM variant with a dynamic forgetting mechanism.

\subsubsection{Review of FP-ELM}
This section provides a brief review of FP-ELM.
The training algorithm of FP-ELM is formulated as follows.
\begin{equation}\label{eq:fp_elm_train}
    \begin{split}
        \bm{K}_i &= \alpha^2_i\bm{K}_{i-1} + \bm{H}^T_i\bm{H}_i \\
        \bm{\beta}_i &= \bm{\beta}_{i-1} +  (\lambda\bm{I} + \bm{K}_i)^{-1} \\
        &\cdot (\bm{H}^T_i(\bm{t}_i - \bm{H}_i\bm{\beta}_{i-1}) - \lambda(1 - \alpha^2_i)\bm{\beta}_{i-1}) \\
    \end{split}
\end{equation}
Especially, $\bm{K}_0$ and $\bm{\beta}_0$ are computed as follows.
\begin{equation}\label{eq:fp_elm_init}
    \begin{split}
        \bm{K}_0 &= \bm{H}_0^T\bm{H}_0 \\
        \bm{\beta}_0 &= (\lambda \bm{I} + \bm{H}_0^T\bm{H}_0)^{-1}\bm{H}_0^T\bm{t}_0,
    \end{split}
\end{equation}
where $\lambda$ is the L2 regularization parameter for $\bm{\beta}$.
$\lambda$ limits $\|\bm{\beta}\|_2$ so that it does not become too large to prevent overfitting.
$0 < \alpha_i \leq 1$ is the forgetting factor that controls
the weight (i.e., the significance) of each past training chunk.
Suppose the latest training step is $i$; then $w_k$, the weight of the $k$th training chunk, is gradually decreased
from one step to the next, as shown below.
\begin{eqnarray}\label{eq:fp_elm_w}
    w_k =\left\{ \begin{array}{ll}
    \prod_{j=k+1}^i \alpha_j, & (0 \leq  k \leq i-1) \\
    1, & (k=i) \\
    \end{array} \right.
\end{eqnarray}
Please note that $\alpha_i$ is a variable parameter that can be adaptively updated according to the information
in the arriving input data or output error values.

\subsubsection{Proposed Forgetting Mechanism}\label{sssec:propose_forget}
FP-ELM can control the weights of past training chunks.
However, it cannot remove the matrix inversion $(\lambda\bm{I} + \bm{P}_i)^{-1}$ in Equation \ref{eq:fp_elm_train} even when the batch equals 1,
because the size of the target matrix $\lambda\bm{I} + \bm{P}_i$ is $\tilde{N} \times \tilde{N}$, where $\tilde{N}$ denotes the number of hidden nodes.
To address this issue, we modify FP-ELM so that it can remove the matrix inversion when batch size = 1.

First, the following equations are derived by disabling the L2 regularization trick (i.e., let $\lambda$ = 0) in Equation \ref{eq:fp_elm_train}.
\begin{equation}\label{eq:fp_elm_no_reg}
    \begin{split}
        \bm{K}_i &= \alpha^2_i \bm{K}_{i-1} + \bm{H}^T_i \bm{H}_i \\
        \bm{\beta}_i &= \bm{\beta}_{i-1} +  \bm{K}_i^{-1}\bm{H}^T_i (\bm{t}_i - \bm{H}_i \bm{\beta}_{i-1})
    \end{split}
\end{equation}
Next, the update formula of $\bm{K}_i^{-1}$ is derived with the Woodbury formula \cite{woodbury}
\footnote{$(\bm{A} + \bm{U}\bm{C}\bm{V})^{-1} = \bm{A}^{-1} - \bm{A}^{-1}\bm{U}(\bm{C}^{-1} + \bm{V}\bm{A}^{-1}\bm{U})^{-1}\bm{V}\bm{A}^{-1}$}.
\begin{equation}\label{eq:fp_elm_woodbury}
    \begin{split}
        \bm{K}_i^{-1} &= (\alpha^2_i \bm{K}_{i-1} + \bm{H}^T_i \bm{H}_i)^{-1} \\
        &= (\frac{1}{\alpha^2_i}\bm{K}_{i-1}^{-1}) - (\frac{1}{\alpha^2_i}\bm{K}_{i-1}^{-1})\bm{H}^T_i \\
        &\cdot (\bm{I} + \bm{H}_i (\frac{1}{\alpha^2_i} \bm{K}_{i-1}^{-1}) \bm{H}^T_i)^{-1} \bm{H}_i (\frac{1}{\alpha^2_i} \bm{K}_{i-1}^{-1})
    \end{split}
\end{equation}
Finally, the training algorithm is obtained by defining $\bm{P}_i \equiv \bm{K}_i^{-1}$.
\begin{equation}\label{eq:lfp_elm_train}
    \begin{split}
        \bm{P}_i &= (\frac{1}{\alpha^2_i}\bm{P}_{i-1}) - (\frac{1}{\alpha^2_i}\bm{P}_{i-1})\bm{H}^T_i \\
        &\cdot (\bm{I} + \bm{H}_i (\frac{1}{\alpha^2_i} \bm{P}_{i-1}) \bm{H}^T_i)^{-1} \bm{H}_i (\frac{1}{\alpha^2_i} \bm{P}_{i-1}) \\
        \bm{\beta}_i &= \bm{\beta}_{i-1} + \bm{P}_i\bm{H}^T_i(\bm{t}_i - \bm{H}_i\bm{\beta}_{i-1})
    \end{split}
\end{equation}
$\bm{P}_0$ and $\bm{\beta}_0$ are computed with the same algorithm as Equation \ref{eq:os_elm_init}.
The proposed forgetting mechanism eliminates the matrix inversion in Equation \ref{eq:lfp_elm_train} when batch size = 1 because
the size of the target matrix $\bm{I} + \bm{H}_i (\frac{1}{\alpha^2_i} \bm{P}_{i-1}) \bm{H}^T_i$ is $k \times k$, where $k$ denotes the batch size.
Equation \ref{eq:lfp_elm_train} becomes equal to the original training algorithm of OS-ELM when $\frac{1}{\alpha_i^2}\bm{P}_i$ is replaced with $\bm{P}_i$.
Thus, the proposed method provides a forgetting function with a tiny additional computational cost
to the original training algorithm of OS-ELM.
However, it may suffer from overfitting, since the L2 regularization trick is disabled.
The trade-off is quantitatively evaluated in Section \ref{sec:eval_anomaly}.

\subsection{Algorithm}\label{ssec:detector_algo}
{\detector} leverages OS-ELM of batch size = 1 in conjunction with the proposed forgetting mechanism.
The following equations are derived by combining Equations \ref{eq:os_elm_train_bsize_1} and \ref{eq:lfp_elm_train}.
\begin{eqnarray}\label{eq:lfp_elm_train_bsize_1}
    \begin{split}
        \bm{P}_i &= (\frac{1}{\alpha_i^2}\bm{P}_{i-1}) - \frac{(\frac{1}{\alpha_i^2}\bm{P}_{i-1})\bm{h}_i^T\bm{h}_i(\frac{1}{\alpha_i^2}\bm{P}_{i-1})}{1+\bm{h}_i(\frac{1}{\alpha_i^2}\bm{P}_{i-1})\bm{h}_i^T} \\
        \bm{\beta}_i &= \bm{\beta}_{i-1} + \bm{P}_i\bm{h}_i^T(\bm{t}_i - \bm{h}_i\bm{\beta}_{i-1})
    \end{split}
\end{eqnarray}
{\detector} is built on an OS-ELM-based autoencoder to construct a semi-supervised anomaly detector; $\bm{t}_i = \bm{x}_i$ holds in Equation \ref{eq:lfp_elm_train_bsize_1}.
The training algorithm of {\detector} is as follows.
\begin{eqnarray}\label{eq:detector_train}
    \begin{split}
        \bm{P}_i &= (\frac{1}{\alpha_i^2}\bm{P}_{i-1}) - \frac{(\frac{1}{\alpha_i^2}\bm{P}_{i-1})\bm{h}_i^T\bm{h}_i(\frac{1}{\alpha_i^2}\bm{P}_{i-1})}{1+\bm{h}_i(\frac{1}{\alpha_i^2}\bm{P}_{i-1})\bm{h}_i^T} \\
        \bm{\beta}_i &= \bm{\beta}_{i-1} + \bm{P}_i\bm{h}_i^T(\bm{x}_i - \bm{h}_i\bm{\beta}_{i-1})
    \end{split}
\end{eqnarray}
$\bm{P}_0$ and $\bm{\beta}_0$ are computed as follows (there are no changes from Equation \ref{eq:os_elm_init}).
\begin{equation}\label{eq:detector_init}
    \begin{split}
        \bm{P}_0 &= (\bm{H}_0^T\bm{H}_0)^{-1}\\
        \bm{\beta}_0 &= \bm{P}_0\bm{H}_0^T\bm{t}_0
    \end{split}
\end{equation}
As indicated in Equation \ref{eq:detector_train}, {\detector} performs training and forgetting operations at the same time.

The prediction algorithm is formulated as follows.
\begin{equation}\label{eq:detector_predict}
    score = L(\bm{x}, G(\bm{x} \cdot \bm{\alpha} + \bm{b})\bm{\beta}),
\end{equation}
where $L$ denotes a loss function, and $score$ is an anomaly score of $\bm{x}$.

\subsection{Stability of OS-ELM Training}\label{ssec:detector_stab}
OS-ELM has a training stability issue: if $\bm{I} + \bm{H}_i\bm{P}_{i-1}\bm{H}_i^T$ in Equation \ref{eq:os_elm_train} is close to a singular matrix,
the training becomes unstable regardless of the batch size \cite{os_elm}.
In the context of {\detector}, the problem occurs when $1+\bm{h}_i(\frac{1}{\alpha_i^2}\bm{P}_{i-1})\bm{h}_i^T$ in Equation \ref{eq:detector_train} is close to 0.
Thus, {\detector} should stop the training when $\epsilon > 1+\bm{h}_i(\frac{1}{\alpha_i^2}\bm{P}_{i-1})\bm{h}_i^T$, where
$\epsilon$ denotes a small positive value.

\subsection{Example of {\detector} in Practical Use}
The following is an example of {\detector} (shown in Algorithm \ref{alg:detector}) intended for practical use.
First, $\bm{\alpha}$ and $\bm{b}$ are initialized with random values;
then $\bm{\beta}_0$ and $\bm{P}_0$ are computed with Equation \ref{eq:detector_init}.
Please note that the number of initial training samples $k_0$ should be larger
than that of hidden nodes $\tilde{N}$ to make $\bm{H}_0^T\bm{H}_0$ nonsingular.
At the $i$th training step in the following loop, the inequality $\epsilon > 1+\bm{h}_i(\frac{1}{\alpha_i^2}\bm{P}_{i-1})\bm{h}_i^T$ is evaluated,
then the rest of the lines are skipped if it is true.
If it is false, then an anomaly score of $\bm{x}_i$ is computed with Equation \ref{eq:detector_predict}.
$\bm{x}_i$ is judged to be an anomaly if the score is greater than a user-defined threshold $\theta$;
otherwise {\detector} judges $\bm{x}_i$ to be a normal sample.
Finally, sequential learning is performed with Equation \ref{eq:detector_train}.
\begin{algorithm}[t]
    \footnotesize
    \caption{Example of Using {\detector}}
    \begin{algorithmic}[1]\label{alg:detector}
    \STATE{$\bm{\alpha} \leftarrow {\rm random}()$, $\bm{b} \leftarrow {\rm random}()$}
    \STATE{$\bm{H}_0 \leftarrow G(\bm{x_0} \in \bm{R}^{k_0 \times n} \cdot\bm{\alpha} + \bm{b})$} \COMMENT{$k_0 \gg \tilde{N}$}
    \STATE{$\bm{P}_0 \leftarrow (\bm{H}_0^T\bm{H}_0)^{-1}$, $\bm{\beta}_0 \leftarrow \bm{P}_0\bm{H}_0^T\bm{t}_0$}
    \STATE{$i \leftarrow 1$}
    \FOR{until $\{\bm{x}_i \in \bm{R}^n, 0 < \alpha_i \leq 1\}$ exists}
        \STATE{$\bm{h}_i \leftarrow G(\bm{x}_i \cdot \bm{\alpha} + \bm{b})$}
        \IF{$\epsilon > 1 + \bm{h}_i(\frac{1}{\alpha_i^2}\bm{P}_{i-1})\bm{h}_i^T$}
            \STATE{${\rm print}$(``Singular matrix encountered.'')}
            \STATE{$i \leftarrow i + 1$}
            \STATE{continue}
        \ENDIF
        \STATE{$score \leftarrow L(\bm{x}_i, \bm{h}_i\bm{\beta}_{i-1})$}
        \IF{$score > \theta$}
            \STATE{${\rm print}$(``Anomaly detected.'')}
        \ENDIF
        \STATE{$\bm{P}_{i-1} \leftarrow \frac{1}{\alpha_i^2}\bm{P}_{i-1}$}
        \STATE{$\bm{P}_i \leftarrow \bm{P}_{i-1} - \frac{\bm{P}_{i-1}\bm{h}_i^T\bm{h}_i\bm{P}_{i-1}}{1+\bm{h}_i\bm{P}_{i-1}\bm{h}_i^T}$}
        \STATE{$\bm{\beta}_i \leftarrow \bm{\beta}_{i-1} + \bm{P}_i\bm{h}_i^T(\bm{x}_i - \bm{h}_i\bm{\beta}_{i-1})$}
        \STATE{$i \leftarrow i + 1$}
    \ENDFOR
    \end{algorithmic}
\end{algorithm}
\begin{figure}[t]
    \begin{minipage}{0.5\textwidth}
        \centering
        \includegraphics[width=57.5mm]{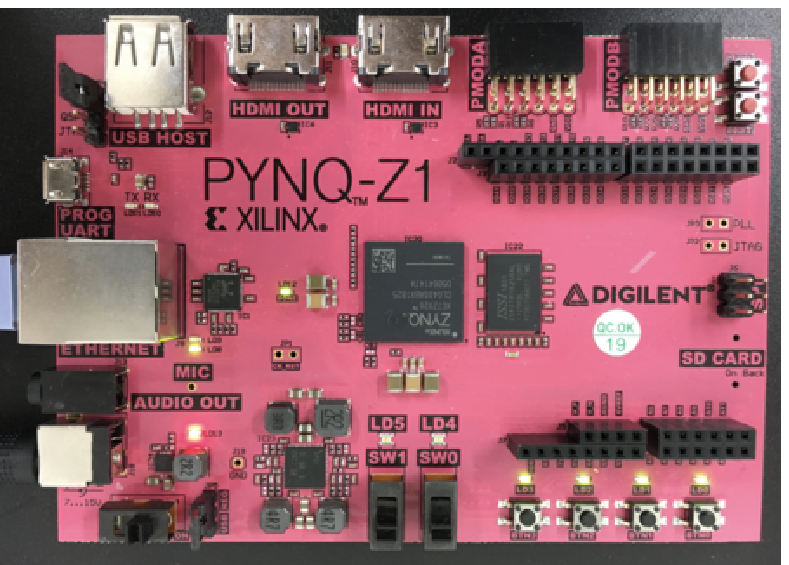}
        \caption{PYNQ-Z1 Board}
        \label{fig:board}
    \end{minipage}
    \begin{minipage}{0.5\textwidth}
        \centering
        \makeatletter
        \def\@captype{table}
        \caption{Specifications of PYNQ-Z1 Board}
        \makeatother
        \footnotesize
        \begin{tabular}{c|c} \hline\hline
            \multicolumn{2}{c}{Board Specifications} \\\hline
            Linux Image & PYNQ v2.4 (Ubuntu v18.04)\\
            SoC Chip & \begin{tabular}{c}Xilinx ZYNQ XC7Z020-1CLG400C\\ CPU: ARM Cortext-A9 650MHz \\ FPGA: Artix-7 \end{tabular} \\
            DRAM & DDR3 512MB \\\hline
            \multicolumn{2}{c}{FPGA Specifications} \\\hline
            BRAM & 280 blocks \\
            DSP & 220 slices \\
            FF & 106,400 instances \\
            LUT & 53,200 instances \\\hline
        \end{tabular}
        \label{tb:board}
    \end{minipage}
\end{figure}
\section{{\core}}\label{sec:imple}
This section describes the design and implementation of {\core}, an IP core of {\detector}.
To demonstrate that {\core} can be implemented on edge devices with limited resources,
we use PYNQ-Z1 board, a low-cost SoC platform where an FPGA is integrated.
Figure \ref{fig:board} displays the board, and its specifications are shown in Table \ref{tb:board}.
We develop {\core} with Vivado HLS v2018.3
and implement it on PYNQ-Z1 board using Vivado v2018.3.
The clock frequency of {\core} is set to 100.0 MHz.

\subsection{Overview of Board-level Implementation}\label{ssec:imple_overview}
\begin{figure}[t]
    \begin{minipage}{0.5\textwidth}
        \centering
        \includegraphics[width=90.0mm]{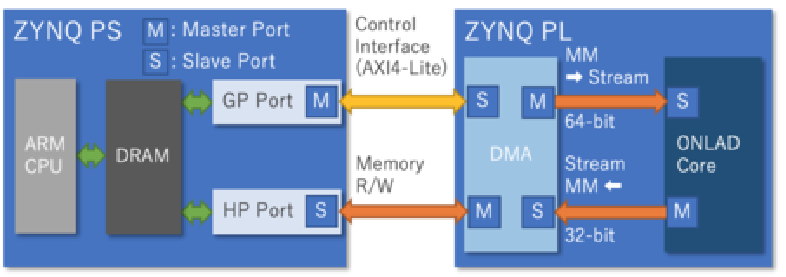}
        \caption{Block Diagram of Board-level Implementation}
        \label{fig:overview}
    \end{minipage}
    \begin{minipage}{0.5\textwidth}
        \centering
        \includegraphics[width=57.5mm]{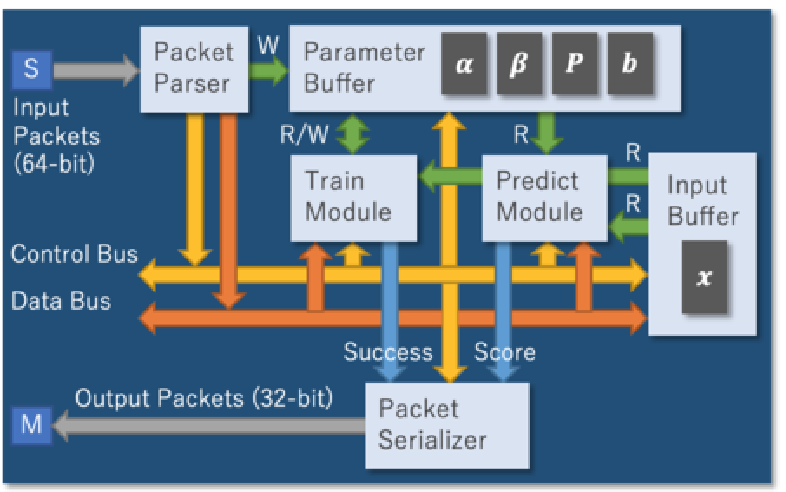}
        \caption{Block Diagram of {\core}}
        \label{fig:osuad_core}
    \end{minipage}
\end{figure}

First, we provide a brief overview of our board-level implementation.
Figure \ref{fig:overview} shows the block diagram.
The PS (Processing System) part is mainly responsible for preprocessing of input data and triggering a DMA (Direct Memory Access) controller.
The DMA controller converts preprocessed input data in DRAM to AXI4-Stream format packets and transfers them to {\core}.
It also converts output packets of {\core} back to AXI4-Memory-Mapped format data, and transfers them to DRAM.
On the other hand, the PL (Programmable Logic) part implements {\core}.
{\core} performs training or prediction computations according to the information in the header of input packets (the details are to be described later).

\subsection{Details of {\core}}\label{ssec:imple_osuad_core}
Figure \ref{fig:osuad_core} illustrates the block diagram of {\core},
with its four important sub-modules: (1) Parameter Buffer, (2) Input Buffer, (3) Train Module, and (4) Predict Module.
The rest of this section explains these sub-modules one by one.

\subsubsection{Parameter Buffer}\label{sssec:parameter_buffer}
Parameter Buffer manages the parameters of {\core} (i.e., $\bm{\alpha}$, $\bm{\beta}$, $\bm{P}$, and $\bm{b}$).
All the parameters are implemented with BRAMs; hence, more BRAM instances are consumed as the sizes of the parameters increase.
Specifically, the total number of matrix elements of Parameter Buffer (denoted as $S_{param}$) is calculated as follows.
\begin{equation}\label{eq:param_space}
    S_{parameter} = \tilde{N}^2 + (2n + 1)\tilde{N}
\end{equation}
Please note that $n = m$ is applied in Equation \ref{eq:param_space}, since {\detector} is an autoencoder.
Equation \ref{eq:param_space} shows that the utilization of the BRAM instances of this module is proportional to the square of the number of hidden nodes $\tilde{N}^2$
and is also proportional to the number of input nodes $n$.

\subsubsection{Input Buffer}
Input Buffer stores a single input vector preprocessed in the PS part and, like Parameter Buffer, is implemented with BRAMs.
The total number of matrix elements of Input Buffer (denoted as $S_{input}$) is calculated as follows.
\begin{equation}\label{eq:input_space}
    S_{input} = n
\end{equation}
Equation \ref{eq:input_space} shows that the utilization of the BRAM instances of this module is proportional to the number of input nodes $n$.
This module is shared with Train Module and Predict Module so that they can read the input vector.

\subsubsection{Train Module}\label{sssec:train_module}
\begin{figure}
    \begin{minipage}{0.5\textwidth}
        \centering
        \includegraphics[height=55.0mm]{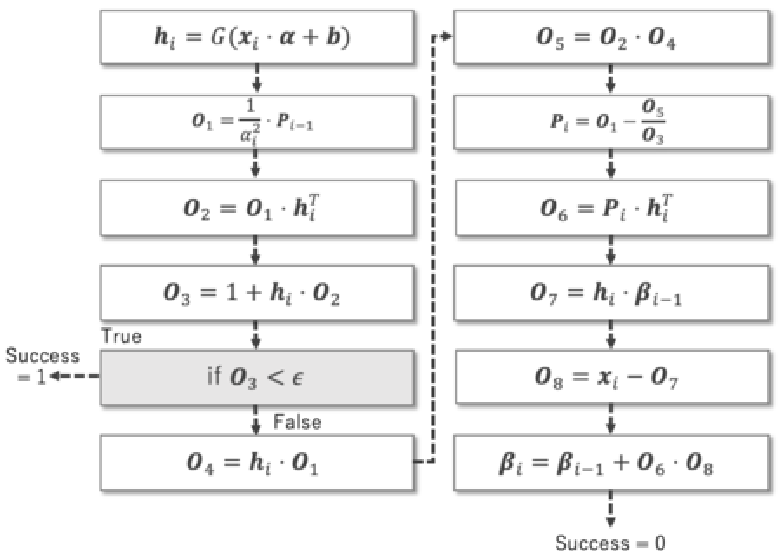}
        \caption{Processing Flow of Train Module}
        \label{fig:train_module}
    \end{minipage}
    \begin{minipage}{0.5\textwidth}
        \centering
        \includegraphics[height=55.0mm]{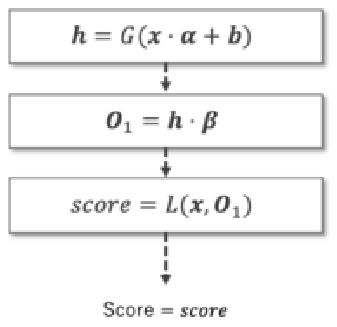}
        \caption{Processing Flow of Predict Module}
        \label{fig:predict_module}
    \end{minipage}
\end{figure}

Train Module executes the training algorithm (i.e., Equation \ref{eq:detector_train}) in order to update the parameters in Parameter Buffer.
Figure \ref{fig:train_module} shows the processing flow.
Each processing block is sequentially executed.
According to the discussion in Section \ref{ssec:detector_stab},
Train Module is designed to interrupt the computation
when $\bm{O}_3 < \epsilon$ holds.
In our implementation, $\epsilon$ is set to $1e^{-4}$.
The output signal of ${\rm Success}$ indicates whether the inequality is satisfied or not (1/0 means satisfied/not satisfied).
All the matrix operations, including matrix products, matrix adds, matrix subs, and element-wise multiplies are implemented with arithmetic units of 32-bit fixed-point precision using DSPs.
To save hardware resources, these matrix operations are designed to use a specific number of arithmetic units regardless of the number of input and hidden nodes.
The matrices shown in the processing flow (i.e., $\bm{O}_{1\sim8}$ and $\bm{h}_i$) are implemented with BRAMs.

The total number of matrix elements of Train Module (denoted as $S_{train}$) is calculated as follows.
\begin{equation}\label{eq:train_space}
    S_{train} = 2\tilde{N}^2 + 4\tilde{N} + 2n + 1
\end{equation}
Equation \ref{eq:train_space} shows that the utilization of the BRAM instances of this module is proportional to the
square of the number of hidden nodes $\tilde{N}^2$
and is also proportional to the number of input nodes $n$.

$I_{train}$ below denotes the total computational iterations needed to finish the processing flow, calculated in the manner described in Section \ref{ssec:detector_anal}.
\begin{equation}\label{eq:train_iter}
    I_{train} = 4\tilde{N}^2 + (3n + 1)\tilde{N}
\end{equation}
The computational cost is proportional to the square of the number of hidden nodes $\tilde{N}^2$,
and is also proportional to the number of input nodes $n$.

\subsubsection{Predict Module}\label{sssec:predict_module}
Predict Module executes the prediction algorithm (i.e., Equation \ref{eq:detector_predict}) to output anomaly scores.
Figure \ref{fig:predict_module} shows the processing flow.
Predict Module follows the design methodology of Train Module.

The total number of matrix elements of Predict Module (denoted as $S_{predict}$) is calculated as follows.
\begin{equation}\label{eq:predict_space}
    S_{predict} = \tilde{N} + n
\end{equation}
Equation \ref{eq:predict_space} shows that the utilization of the BRAM instances of this module is proportional to
the numbers of hidden nodes $\tilde{N}$ and input nodes $n$.

$I_{predict}$ below denotes the total computational iterations to finish the processing flow.
\begin{equation}\label{eq:predict_iter}
    I_{predict} = 2n\tilde{N}
\end{equation}
The computational cost is proportional to the numbers of hidden nodes $\tilde{N}$ and input nodes $n$.

\subsubsection{Implementation of Matrix Operations}
In ONLAD Core, matrix operations, such as matrix product,
matrix add, matrix sub, and element-wise multiply,
are implemented as a dedicated circuit.
These matrix operations are designed with C-level language and
synthesized with Vivado HLS.
Loop unrolling and loop pipelining directives are used
in the innermost loops of these operations for parallelization.
In this design, unrolling factor is set to 2,
so that they are parallelized with two arithmetic units.

\subsection{Instructions of {\core}}
{\core} is designed to execute the following instructions: (1) {\it update\_params}, (2) {\it update\_input}, (3) {\it update\_ff}, (4) {\it do\_training}, and (5) {\it do\_prediction}.
The packet format of each instruction is detailed in Figure \ref{fig:packet_format}.
An input packet is of 64 bits long.
The first 3-bit field (i.e., Mode Field) specifies an instruction to be executed on {\core}.
The following 61-bit field (i.e., Data Field) is reserved for several uses according to the instruction.
An output packet is of 32 bits long and embeds an output result of Train Module or Predict Module.

In the rest of this section, we describe how the sub-modules of {\core} work according to each instruction.

\subsubsection{{\it update\_params}}
\begin{figure}[t]
    \centering
    \includegraphics[width=85.0mm]{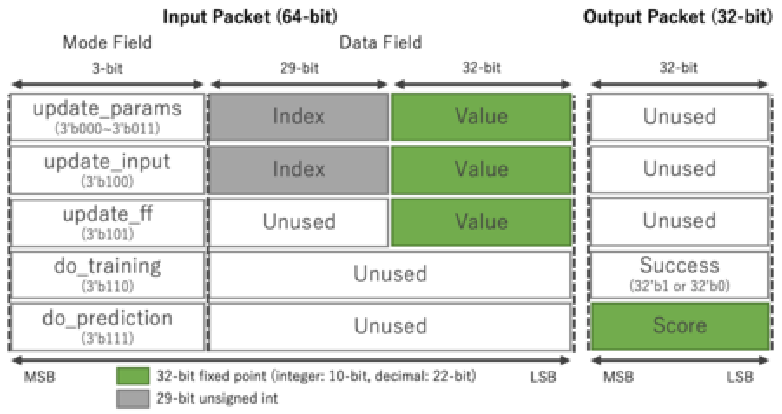}
    \caption{Packet Formats}
    \label{fig:packet_format}
\end{figure}

This instruction updates Parameter Buffer.
The packet format is shown in the first row of Figure \ref{fig:packet_format}.
The target parameter is specified in Mode Field of an input packet.
${\rm Index}$ in Data Field embeds an index of the target parameter, and ${\rm Value}$ an update value.
The target parameter is updated as below.
\begin{equation}\label{eq:osuad_core_init}
    \bm{target}[{\rm Index}] \leftarrow {\rm Value}
\end{equation}
Please note that all the parameters are managed as row-major flattened 1-D arrays in Parameter Buffer.

\subsubsection{{\it update\_input}}
This instruction updates Input Buffer.
The packet format (the second row of Figure \ref{fig:packet_format}) is almost the same as {\it update\_params} instruction except for Mode Field.
\begin{equation}\label{eq:osuad_core_input}
    \bm{x}[{\rm Index}] \leftarrow {\rm Value}
\end{equation}
Input Buffer is updated with the above formula.
Please note that $n$ input packets are required to create an $n$-dimensional input vector.

\subsubsection{{\it update\_ff}}
This instruction updates the forgetting factor $\alpha_i$ managed in Train Module.
The packet format of this instruction is shown in the third row of Figure \ref{fig:packet_format}.
${\rm Value}$ in Data Field embeds an update value.
$\alpha_i$ is updated as follows.
\begin{equation}\label{eq:osuad_core_ff}
    \alpha_i \leftarrow {\rm Value}
\end{equation}

\subsubsection{{\it do\_training}}\label{sssec:sequential_learning}
This instruction executes training computations with Train Module.
Train Module first reads the latest parameters (i.e., $\bm{\beta}_{i-1}$ and $\bm{P}_{i-1}$) from Parameter Buffer, and an input vector from Input Buffer.
Then, it executes the training algorithm and
updates Parameter Buffer with the new parameters (i.e., $\bm{\beta}_{i}$ and $\bm{P}_{i}$).

The packet format is shown in the fourth row of Figure \ref{fig:packet_format}.
An input packet of this instruction is just a trigger to perform training.
An output packet of this instruction embeds an evaluation result (denoted as ${\rm Success}$) of the inequality described in Section \ref{sssec:train_module} (1/0 means satisfied/not satisfied).

\begin{table}[b]
    \centering
    \footnotesize
    \caption{Datasets}
    \label{tb:dataset}
    \begin{tabular}{c|c|c|c}\hline\hline
    Name & Samples & Features & Classes\\\hline
    Fashion MNIST \cite{fashion_mnist} & 70,000 & 784 & 10\\
    MNIST \cite{mnist} & 70,000 & 784 & 10\\
    Smartphone HAR \cite{uciml} & 5,744 & 561 & 6\\
    Drive Diagnosis \cite{uciml} & 58,509 & 48 & 11\\
    Letter Recognition \cite{uciml} & 20,000 & 16 & 26\\\hline
    \end{tabular}
\end{table}
\subsubsection{{\it do\_prediction}}
This instruction executes prediction computations with Predict Module.
Predict Module reads the latest output weight if it is updated,
and an input vector in the same way as Train Module.
Predict Module then executes the prediction algorithm and outputs an anomaly score of the input vector.

The packet format is shown in the last row of Figure \ref{fig:packet_format}.
An input packet of this instruction is also just a trigger for prediction.
An output packet of this instruction embeds an output anomaly score (denoted as ${\rm Score}$) computed by {\core}.

\section{Evaluation of {\detector}}\label{sec:eval_anomaly}
In this section, the anomaly detection capability of {\detector} is evaluated in comparison with other models.
A common server machine (OS: Ubuntu 18.04, CPU: Intel Core i7 6700 3.4GHz, GPU: Nvidia GTX 1070 8GB, DRAM: DDR4 16GB, and Storage: SSD 512GB)
is used as the experimental machine in this section and Section \ref{sec:eval_hard}.
\begin{table}[b]
    \centering
    \footnotesize
    \caption{Search Ranges of Hyperparameters}
    \label{tb:exploration}
    \begin{tabular}{c|c|c} \hline\hline
    & {\detector} & {\fpelm} \\\hline    
    $G_{hidden}$ & \{Sigmoid \cite{Sigmoid}, Identity\footnotemark\} & \{Sigmoid, Identity\} \\
    $p(x)$ & Uniform [0,1] &  Uniform [0,1] \\
    $L$ & MSE\footnotemark & MSE \\
    $\alpha_i$ & \{0.95, 0.96, \dots, 1.00\} & \{0.95, 0.96, \dots, 1.00\} \\
    $\tilde{N}_1$ & \{8, 16, 32, \dots, 256\} & \{8, 16, 32, \dots, 256\} \\
    $\lambda$ & & 0.02\footnotemark \\\hline
    & {\nn} & {\dnn} \\\hline
    $G_{hidden}$ & \{Sigmoid, Relu \cite{relu}\} & \{Sigmoid, Relu\} \\
    $G_{out}$ & Sigmoid &  Sigmoid \\
    $L$ & MSE & MSE \\
    $O$ & Adam \cite{adam} & Adam \\
    $B$ & \{8, 16, 32\} & \{8, 16, 32\} \\
    $E$ & \{5, 10, 15, 20\} & \{5, 10, 15, 20\} \\
    $\tilde{N}_1$ & \{8, 16, 32, \dots, 256\} & \{8, 16, 32, \dots, 256\} \\
    $\tilde{N}_2$ & & \{8, 16, 32, \dots, 256\} \\
    $\tilde{N}_3$ & & \{8, 16, 32, \dots, 256\} \\\hline
    \end{tabular}
\end{table}

\subsection{Experimental Setup}\label{ssec:eval_setup}
{\detector} is compared with the following models: (1) \textbf{{\fpelm}}, (2) \textbf{{\nn}}, and (3) \textbf{{\dnn}}.
{\fpelm} is an FP-ELM-based autoencoder.
This model is used to quantitatively evaluate the effect of disabling the L2 regularization trick in {\detector}.
{\nn} is a 3-layer BP-NN-based autoencoder, and {\dnn} is a BP-NN-based deep autoencoder consisting of five layers.
These models are used to compare OS-ELM-based autoencoders (i.e., {\fpelm} and {\detector}) with BP-NN-based ones.
All the models, including {\detector}, were implemented with TensorFlow v1.13.1 \cite{tensorflow}.

For a comprehensive evaluation, two testbeds: (1) \textbf{Offline Testbed} and (2) \textbf{Online Testbed} are conducted.
Offline Testbed simulates an environment where all training and test data are available in advance and no concept drift occurs.
This is a standard experimental setup to evaluate semi-supervised anomaly detection models.
The purpose of Offline Testbed is to measure the generalization capability of {\detector} in the context of anomaly detection.
This testbed is not used to evaluate the proposed forgetting mechanism (i.e., $\alpha_i$ is always fixed to 1), since no concept drift occurs in this testbed.
On the other hand, Online Testbed simulates an environment where at first only a small part of a dataset is given and the rest arrives as time goes by.
Online Testbed assumes that concept drift occurs.
The purpose of this testbed is to evaluate the robustness of the proposed forgetting mechanism against concept drift in comparison with the other models.

Several public classification datasets listed in Table \ref{tb:dataset} are used to construct Offline Testbed and Online Testbed.
All data samples are normalized within [0,1] by using min-max normalization.
Hyperparameters of each model are explored within the ranges detailed in Table \ref{tb:exploration}\footnote{
    $G_{hidden}$: an activation function applied to all the hidden layers.
    $G_{out}$: an activation function applied to the output layer.
    $p(x)$: a probability density function used for random initialization of {\detector} and {\fpelm}.
    $\tilde{N}_{i}$: the number of nodes of the $i$th hidden layer.
    $L$: a loss function.
    $\alpha_i$: the forgetting factor of {\detector} and {\fpelm}.
    $\lambda$: the L2 regularization parameter of {\fpelm}.
    $O$: an optimization algorithm.
    $B$: batch size.
    $E$: the number of training epochs.
}.
\footnotetext[4]{$G(\bm{x}) = \bm{x}$}
\footnotetext[5]{$L(\bm{x}, \bm{y}) = \frac{1}{n}\sum_{i=0}^{n}(x_i - y_i)^2$}
\footnotetext[6]{This value was used for the experiments in the original paper of FP-ELM (i.e., \cite{fp_elm}).}

\subsection{Experimental Method}\label{ssec:eval_anomaly_method}
This section describes the experimental methods of Offline Testbed and Online Testbed, respectively.

Algorithm \ref{alg:eval_offline} shows the experimental method of Offline Testbed.
In this testbed, a dataset is divided into training data $\bm{X}_{train}$ (80\%) and test data $\bm{X}_{test}$ (20\%), respectively.
Suppose we have a dataset that consists of $c$ classes in total;
training data of class $i$ are used as normal data for training (denoted as $\bm{X}_{normal\_train}$)
and test data of class $i$ are as normal data for testing (denoted as $\bm{X}_{normal\_test}$).
Test data of class $j \neq i$ are used as anomaly data (denoted as $\bm{X}_{anomaly}$).
The number of samples in $\bm{X}_{anomaly}$ is limited up to 10\% of that of $\bm{X}_{normal\_test}$
to simulate a practical situation; anomaly data are much rarer than normal data in most cases.
A model is trained with $\bm{X}_{normal\_train}$ ({\nn} and {\dnn} are trained with batch size = $B$ for $E$ epochs).
Once the training procedure is finished,
the model is evaluated with a test set that mixes $\bm{X}_{normal\_test}$ and $\bm{X}_{anomaly}$,
then an AUC (Area Under Curve) score is calculated.
AUC is one of the most widely used metrics for evaluating the accuracy of anomaly detection models independently of particular anomaly score thresholds.
The above process is repeated until $i < c$, then all the $c$ AUC scores are averaged.
The output score is recorded as a result of a single trial;
the final AUC scores reported in Table \ref{tb:eval_offline} are averages over 50 trials.
10-fold cross-validation is conducted for hyperparameter tuning.
\begin{algorithm}[t!]
    \footnotesize
    \caption{Offline Testbed}
    \begin{algorithmic}[1]\label{alg:eval_offline}
    \STATE{$\bm{X}_{train} \equiv [\bm{X}_{train}^{(0)}, \bm{X}_{train}^{(1)}, \dots, \bm{X}_{train}^{(c-1)}]$}
    \STATE{$\bm{X}_{test} \equiv [\bm{X}_{test}^{(0)}, \bm{X}_{test}^{(1)}, \dots, \bm{X}_{test}^{(c-1)}]$}
    \STATE{$average\_auc \leftarrow 0$}
    \FOR{$i \leftarrow 0$ to $c - 1$}
        \STATE{$\bm{X}_{normal\_train} \leftarrow \bm{X}_{train}^{(i)}$}
        \STATE{$\bm{X}_{normal\_test} \leftarrow \bm{X}_{test}^{(i)}$}
        \STATE{$\bm{X}_{anomaly} \leftarrow \bm{X}_{test}^{(j \neq i)}$}
        \STATE{$num\_anomalies \leftarrow {\rm len}(\bm{X}_{normal\_test}) \times 0.1$}
        \STATE{$\bm{X}_{anomaly} \leftarrow {\rm sample}(\bm{X}_{anomaly}, num\_anomalies)$}
        \STATE{$model.{\rm train}(\bm{X}_{normal\_train})$}
        \STATE{$scores \leftarrow model.{\rm predict}({\rm concat}([\bm{X}_{normal\_test}, \bm{X}_{anomaly}]))$}
        \STATE{$average\_auc \leftarrow average\_auc + {\rm calc\_auc}(scores)$}
        \STATE{$model.{\rm reset()}$}
    \ENDFOR
    \STATE{$average\_auc \leftarrow \frac{average\_auc}{c}$}
    \end{algorithmic}
\end{algorithm}
\begin{algorithm}[t!]
    \footnotesize
    \caption{Online Testbed}
    \begin{algorithmic}[1]\label{alg:eval_online}
    \STATE{$\bm{X}_{init} \equiv [\bm{X}_{init}^{(0)}, \bm{X}_{init}^{(1)}, \dots, \bm{X}_{init}^{(c-1)}]$}
    \STATE{$\bm{X}_{test} \equiv [\bm{X}_{test}^{(0)}, \bm{X}_{test}^{(1)}, \dots, \bm{X}_{test}^{(c-1)}]$}
    \STATE{$\bm{X}_{normal}, \bm{X}_{anomaly} \leftarrow {\rm split}(\bm{X}_{test},$``9 : 1''$)$}
    \STATE{$indices \leftarrow [0, 1, \dots, c - 1]$}
    \STATE{${\rm shuffle}(indices)$}
    \STATE{$\bm{X}_{concept} \leftarrow []$}
    \FOR{$i \leftarrow$ to $c - 1$}
        \STATE{$concept \leftarrow [\bm{X}_{normal}^{(indices[i])}]$}
        \STATE{$num\_anomalies \leftarrow {\rm len}(\bm{X}_{normal}^{(indices[i])}) \times 0.1$}
        \STATE{$concept.{\rm append}({\rm sample}(\bm{X}_{anomaly}^{(j \neq indices[i])}, num\_anomalies))$}
        \STATE{$\bm{X}_{concept}.{\rm append}({\rm shuffle}({\rm concat}(concept)))$}
    \ENDFOR
    \STATE{}
    \STATE{$model.{\rm train}(\bm{X}_{init}^{(indices[0])})$}
    \STATE{$scores \leftarrow []$}
    \FOR{$i \leftarrow$ to $c - 1$}
        \FORALL{$\bm{x}$ in $\bm{X}_{concept}[i]$}
            \STATE{$score \leftarrow model.{\rm predict}(\bm{x})$}
            \STATE{$scores.{\rm append}(score)$}
            \STATE{$model.{\rm train}(\bm{x})$}
        \ENDFOR
    \ENDFOR
    \STATE{$auc \leftarrow {\rm calc\_auc}(scores)$}
    \end{algorithmic}
\end{algorithm}

Algorithm \ref{alg:eval_online} shows the experimental method of Online Testbed.
In this testbed, a dataset is divided into initial data $\bm{X}_{init}$ (10\%), test data $\bm{X}_{test}$ (45\%), and validation data $\bm{X}_{valid}$ (45\%).
$\bm{X}_{init}$ represents for data samples that exit in the begining.
$\bm{X}_{test}$ and $\bm{X}_{valid}$ represent for data samples that sequentially arrive as time goes by.
$\bm{X}_{test}$ is used to measure the final AUC scores, while $\bm{X}_{valid}$ is only for hyperparameter tuning.
Both are further divided into normal data $\bm{X}_{normal}$ (90\%) and anomaly data $\bm{X}_{anomaly}$ (10\%).
In the first step, a list (denoted as $indices$) consisting of integers 0$\sim$$c-1$ is constructed and randomly shuffled.
The output indicates the normal class of each concept;
e.g., supposing that $indices = [2, 0, 1]$, the normal class of the 0/1/2th concept is 2/0/1.
The $i$th concept $\bm{X}_{concept}[i]$ mixes normal data of class $indices[i]$ and anomaly data of class $j \neq indices[i]$.
The number of anomaly samples per one concept is limited to 10\% of that of normal samples.
A model is trained with initial data of the first normal class $\bm{X}_{init}^{(indices[0])}$({\nn} and {\dnn} are trained with batch size = $B$ for $E$ epochs).
Then, the model computes an anomaly score for each data sample continuously given from $\bm{X}_{concept}[0] \sim \bm{X}_{concept}[c-1]$.
Every time an anomaly score is computed, the model is trained with the data sample
(all the models, including {\nn} and {\dnn}, are trained with batch size = 1 to sequentially follow the transition of the normal class).
After all the data samples are fed to the model, an AUC score is calculated with the anomaly scores.
This AUC score is recorded as a result of a single trial;
the final AUC scores reported in Table \ref{tb:eval_online} are averages over 50 trials.
Hyperparameter tuning is conducted with the same algorithm for 10 trials by replacing $\bm{X}_{test}$ with $\bm{X}_{valid}$ in Algorithm \ref{alg:eval_online}.

\begin{table}[h!]
    \centering
    \footnotesize
    \caption{AUC Scores on Offline Testbed}
    \label{tb:eval_offline}
    \begin{tabular}{c|c|c|c|c}\hline\hline
    Dataset & {\detector} & {\fpelm} & {\nn} & {\dnn} \\\hline
    Fashion MNIST & 0.905 & 0.905 & \textbf{0.925} & 0.913 \\
    MNIST & 0.944 & 0.945 & 0.958 & \textbf{0.961} \\
    Smartphone HAR & \textbf{0.929} & 0.928 & 0.922 & 0.910 \\
    Drive Diagnosis & 0.939 & 0.943 & 0.952 & \textbf{0.961} \\
    Letter Recognition & 0.952 & 0.950 & 0.978 & \textbf{0.985} \\\hline
    \end{tabular}
\end{table}
\begin{table}[h!]
    \centering
    \footnotesize
    \caption{AUC Scores on Online Testbed}
    \label{tb:eval_online}
    \begin{tabular}{c|c|c|c|c|c}\hline\hline
    Dataset & {\noff} & {\detector} & {\fpelm} & {\nn} & {\dnn} \\\hline
    Fashion MNIST & 0.575 & \textbf{0.869} & 0.866 & 0.685 & 0.697 \\
    MNIST & 0.591 & \textbf{0.899} & 0.898 & 0.787 & 0.755 \\
    Smartphone HAR & 0.558 & 0.781 & 0.788 & 0.785 & \textbf{0.799} \\
    Drive Diagnosis & 0.552 & 0.786 & 0.849 & 0.744 & \textbf{0.853} \\
    Letter Recognition & 0.548 & \textbf{0.882} & 0.879 & 0.737 & 0.788 \\\hline
    \end{tabular}
\end{table}
\subsection{Experimental Results}\label{ssec:eval_anomaly_results}
The experimental results for Offline Testbed are shown in Table \ref{tb:eval_offline}.
The hyperparameter settings are also listed in Table \ref{tb:eval_offline_hp}.
Here, {\nn} and {\dnn} achieve slightly higher AUC scores than those of {\detector} by approximately 0.01$\sim$0.03 point on almost all the datasets.
This result implies that BP-NN-based autoencoders have slightly higher generalization capability than that of OS-ELM-based ones in the context of anomaly detection.
However, {\nn} and {\dnn} have to be iteratively trained for some epochs in order to achieve their best performance (here, they were trained for 5$\sim$20 epochs).
In contrast, {\detector} always finds the optimal output weight in only one epoch.
Also, {\detector} achieves its best AUC scores with an equal or smaller size compared with {\nn} and {\dnn} for all the datasets,
which helps to reduce the computational cost and save on hardware resources required to implement {\core}.
In addition, the differences between the AUC scores of {\detector} and {\fpelm} are within 0.001$\sim$0.004 point;
{\detector} keeps favorable generalization performance even when the L2 regularization trick is disabled.
In summary, {\detector} has comparable generalization capability to that of the BP-NN-based models in much smaller training epochs with an equal or smaller model size.

The experimental results for Online Testbed are shown in Table \ref{tb:eval_online}.
The hyperparameter settings are also listed in Table \ref{tb:eval_online_hp}.
Here, another model, named {\noff} ({\detector}-No-Forgetting-mechanism) is introduced in order to examine the effectiveness of the proposed forgetting mechanism.
{\noff} is the special case of {\detector}, where the forgetting mechanism is disabled by setting $\alpha_i$ to 1.
The hyperparameter settings of {\noff} are the same as those of {\detector}, except for $\alpha_i$.
As shown in the table, {\noff} suffers from significantly lower AUC scores than {\detector}.
The reason is quite obvious; {\noff} does not have any functions to forget past learned data,
therefore it gradually becomes more difficult to detect anomalies every time concept drift happens.
{\nn} and {\dnn}, on the other hand, achieve much higher AUC scores than {\noff} because
BP-NNs have the catastrophic forgetting nature \cite{catastrophic_forgetting}, which works as a kind of forgetting mechanism.
However, BP-NNs do not have any numerical parameters to analytically control the progress of forgetting, unlike {\detector}.
For this reason, {\detector} stably achieves more favorable AUC scores.
Additionally, {\detector} and {\fpelm} have similar AUC scores on most of the datasets, as with the results on Offline Testbed.
This result shows that the proposed forgetting mechanism is not significantly affected by the L2 regularization trick on these datasets.
In summary, {\detector} achieves much higher AUC scores than those of {\nn} and {\dnn} by approximately 0.10$\sim$0.18 point on three datasets out of the five ones.
It also achieves comparable AUC scores to those of the BP-NN-based models on the other two datasets.
\begin{table}[t]
    \footnotesize
    \centering
    \caption{Hyperparameter Settings on Offline Testbed}
    \label{tb:eval_offline_hp}
    \begin{tabular}{c|c|c}\hline\hline
    Dataset & \begin{tabular}{c} {\detector} \\ $\{G_{hidden}, p(x), \tilde{N}_1, L, \alpha_i\}$ \end{tabular} & \begin{tabular}{c} {\fpelm} \\ $\{G_{hidden}, p(x), \tilde{N}_1, L, \alpha_i, \lambda\}$ \end{tabular} \\\hline
    Fashion MNIST & $\{$Identity, Uniform, 64, MSE, 1.00$\}$ & $\{$Identity, Uniform, 64, MSE, 1,00, 0.02$\}$ \\
    MNIST & $\{$Identity, Uniform, 64, MSE, 1.00$\}$ & $\{$Identity, Uniform, 64, MSE, 1.00, 0.02$\}$ \\
    Smartphone HAR & $\{$Identity, Uniform, 128, MSE, 1.00$\}$ & $\{$Identity, Uniform, 128, MSE, 1.00, 0.02$\}$ \\
    Drive Diagnosis & $\{$Sigmoid, Uniform, 16, MSE, 1.00$\}$ & $\{$Sigmoid, Uniform, 16, MSE, 1.00, 0.02$\}$ \\
    Letter Recognition & $\{$Sigmoid, Uniform, 8, MSE, 1.00$\}$ & $\{$Sigmoid, Uniform, 8, MSE, 1.00, 0.02$\}$ \\\hline
    Dataset & \begin{tabular}{c} {\nn} \\ $\{G_{hidden}, G_{out}, \tilde{N}_1, L, O, B, E\}$ \end{tabular} & \begin{tabular}{c} {\dnn} \\ $\{G_{hidden}, G_{out}, \tilde{N}_1, \tilde{N}_2, \tilde{N}_3, L, O, B, E\}$ \end{tabular} \\\hline
    Fashion MNIST & $\{$Relu, Sigmoid, 64, MSE, Adam, 32, 5$\}$ & $\{$Relu, Sigmoid, 64, 32, 64, MSE, Adam, 8, 10$\}$ \\
    MNIST & $\{$Relu, Sigmoid, 64, MSE, Adam, 32, 5$\}$ & $\{$Relu, Sigmoid, 64, 32, 64, MSE, Adam, 8, 10$\}$ \\
    Smartphone HAR & $\{$Relu, Sigmoid, 256, MSE, Adam, 8, 20$\}$ & $\{$Relu, Sigmoid, 128, 256, 128, MSE, Adam, 8, 20$\}$ \\
    Drive Diagnosis & $\{$Relu, Sigmoid, 256, MSE, Adam, 8, 10$\}$ & $\{$Relu, Sigmoid, 128, 256, 128, MSE, Adam, 8, 20$\}$ \\
    Letter Recognition & $\{$Relu, Sigmoid, 256, MSE, Adam, 8, 20$\}$ & $\{$Relu, Sigmoid, 128, 256, 128, MSE, Adam, 8, 20$\}$ \\\hline
    \end{tabular}
\end{table}
\begin{table}[t]
    \footnotesize
    \centering
    \caption{Hyperparameter Settings on Online Testbed}
    \label{tb:eval_online_hp}
    \begin{tabular}{c|c|c}\hline\hline
    Dataset & \begin{tabular}{c} {\detector} \\ $\{G_{hidden}, p(x), \tilde{N}_1, L, \alpha_i\}$ \end{tabular} & \begin{tabular}{c} {\fpelm} \\ $\{G_{hidden}, p(x), \tilde{N}_1, L, \alpha_i, \lambda\}$ \end{tabular} \\\hline
    Fashion MNIST & $\{$Sigmoid, Uniform, 64, MSE, 0.99$\}$ & $\{$Sigmoid, Uniform, 64, MSE, 0.99, 0.02$\}$ \\
    MNIST & $\{$Sigmoid, Uniform, 64, MSE, 0.99$\}$ & $\{$Sigmoid, Uniform, 64, MSE, 0.99, 0.02$\}$ \\
    Smartphone HAR & $\{$Identity, Uniform, 16, MSE, 0.97$\}$ & $\{$Sigmoid, Uniform, 16, MSE, 0.97, 0.02$\}$ \\
    Drive Diagnosis & $\{$Sigmoid, Uniform, 16, MSE, 0.99$\}$ & $\{$Sigmoid, Uniform, 16, MSE, 0.97, 0.02$\}$ \\
    Letter Recognition & $\{$Identity, Uniform, 8, MSE, 0.95$\}$ & $\{$Identity, Uniform, 8, MSE, 0.95, 0.02$\}$ \\\hline
    Dataset & \begin{tabular}{c} {\nn} \\ $\{G_{hidden}, G_{out}, \tilde{N}_1, L, O, B, E\}$ \end{tabular} & \begin{tabular}{c} {\dnn} \\ $\{G_{hidden}, G_{out}, \tilde{N}_1, \tilde{N}_2, \tilde{N}_3, L, O, B, E\}$ \end{tabular} \\\hline
    Fashion MNIST & $\{$Relu, Sigmoid, 64, MSE, Adam, 32, 5$\}$ & $\{$Relu, Sigmoid, 64, 32, 64, MSE, Adam, 8, 10$\}$ \\
    MNIST & $\{$Relu, Sigmoid, 64, MSE, Adam, 32, 5$\}$ & $\{$Relu, Sigmoid, 64, 32, 64, MSE, Adam, 8, 10$\}$ \\
    Smartphone HAR & $\{$Sigmoid, Sigmoid, 32, MSE, Adam, 8, 20$\}$ & $\{$Sigmoid, Sigmoid, 32, 2, 32, MSE, Adam, 8, 20$\}$ \\
    Drive Diagnosis & $\{$Sigmoid, Sigmoid, 16, MSE, Adam, 8, 10$\}$ & $\{$Sigmoid, Sigmoid, 16, 8, 16, MSE, Adam, 8, 20$\}$ \\
    Letter Recognition & $\{$Relu, Sigmoid, 16, MSE, Adam, 8, 20$\}$ & $\{$Relu, Sigmoid, 16, 8, 16, MSE, Adam, 8, 20$\}$ \\\hline    
    \end{tabular}
\end{table}
\section{Evaluations of Performance and Cost}\label{sec:eval_hard}
In this section, {\core} is evaluated in terms of latency, FPGA resource utilization, and power consumption
in comparison with software implementations.

\subsection{Experimental Setup}\label{ssec:eval_hard_setup}
{\core} is evaluated in comparison with the following software implementations:
(1) \textbf{{\nncpu}}, (2) \textbf{{\dnncpu}}, (3) \textbf{{\nngpu}}, (4) \textbf{{\dnngpu}}, (5) \textbf{{\fpelmcpu}}, and (6) \textbf{{\fpelmgpu}}.
\{*\}-CPU is executed only with a CPU, while \{*\}-GPU is executed with a GPU in cooperation with a CPU.
All of these implementations are developed with Tensorflow v1.13.1.
Here, Tensorflow v1.13.1 is built with AVX2 (Advanced Vector eXtensions 2) instructions and -O3 option to accelerate CPU computations.
It is also built with CUDA \cite{cuda} v10.0 to enable GPGPU execution.

The hyperparameter settings of the above implementations are detailed in Table \ref{tb:hard_hp}.
$p(x)$, $\alpha_i$, $\lambda$ and $E$ have been omitted from the table because these parameters are unrelated to any of the evaluation metrics (i.e., latency, FPGA resource utilization, and power consumption).
The batch size (i.e., $B$) of {\nn}-\{*\} and {\dnn}-\{*\} is fixed to 1, as with {\core} and {\fpelm}-\{*\} in order to conduct fair comparisons of latency and power consumption.

\subsection{Latency}\label{ssec:eval_hard_latency}
\subsubsection{Training/Prediction Latency}
Here, we refer to ``training latency'' as the elapsed time from
when a model receives an input sample until the training algorithm is computed.
``Prediction latency'' is the elapsed time from when a model receives an
input data sample until an anomaly score is calculated.

\begin{figure}[t!]
    \begin{minipage}{0.33\hsize}
        \centering
        \includegraphics[height=42.5mm]{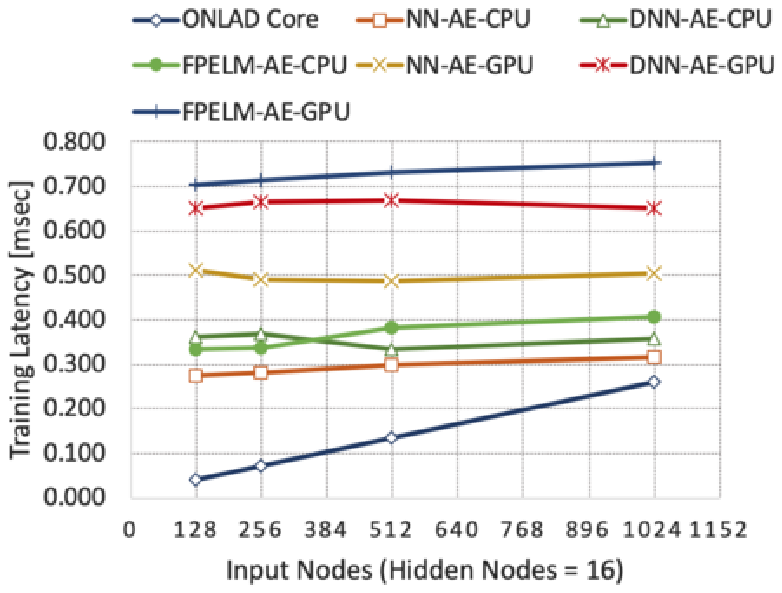}
    \end{minipage}
    \begin{minipage}{0.33\hsize}
        \centering
        \includegraphics[height=42.5mm]{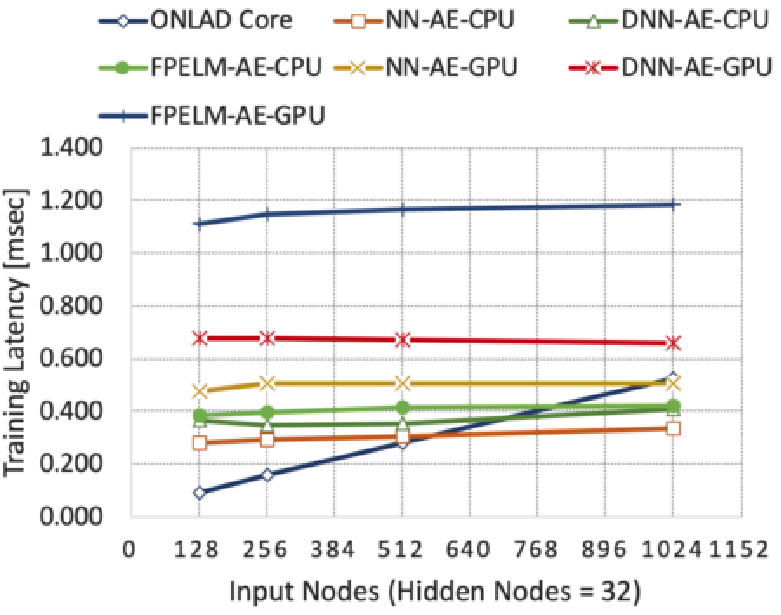}
    \end{minipage}
    \begin{minipage}{0.33\hsize}
        \centering
        \includegraphics[height=42.5mm]{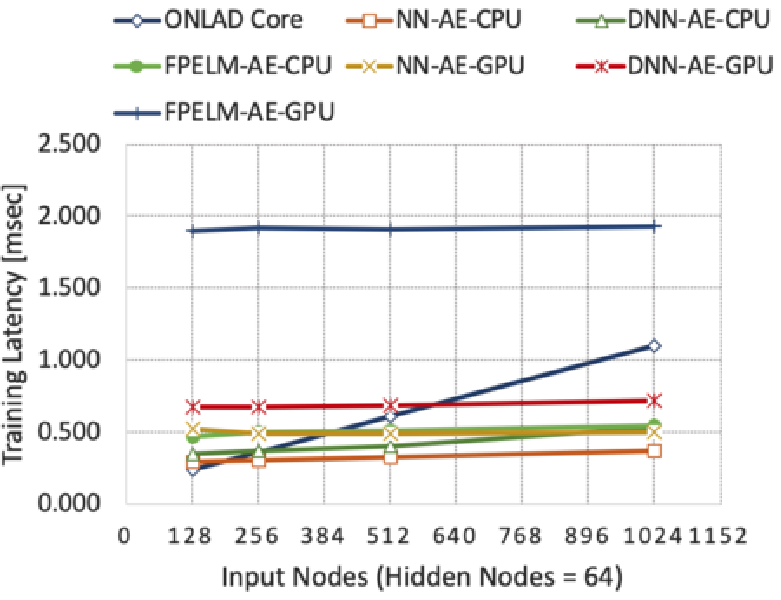}
    \end{minipage}
    \caption{Comparison of Training Latency}
    \label{fig:train_latency}
\end{figure}

\begin{figure}[t!]
    \begin{minipage}{0.33\hsize}
        \centering
        \includegraphics[height=42.5mm]{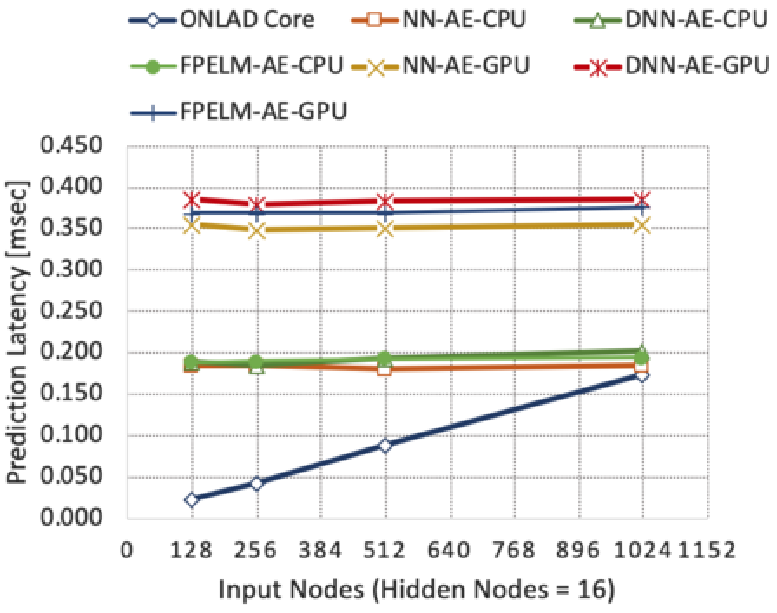}
    \end{minipage}
    \begin{minipage}{0.33\hsize}
        \centering
        \includegraphics[height=42.5mm]{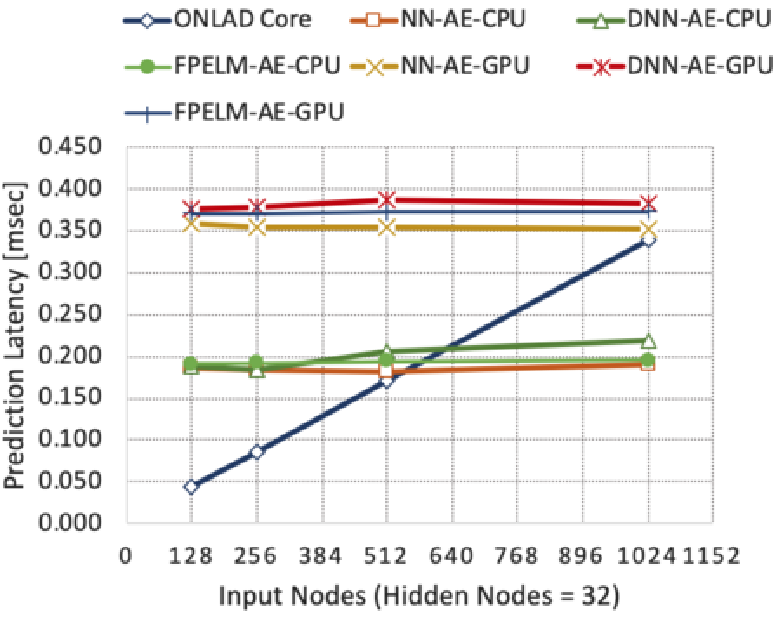}
    \end{minipage}
    \begin{minipage}{0.33\hsize}
        \centering
        \includegraphics[height=42.5mm]{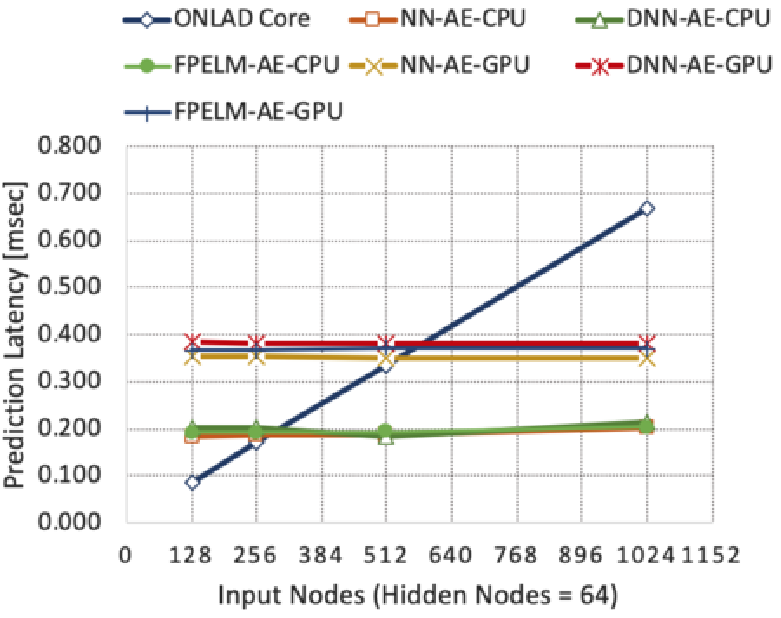}
    \end{minipage}
    \caption{Comparison of Prediction Latency}
    \label{fig:predict_latency}
\end{figure}

Figures \ref{fig:train_latency} and \ref{fig:predict_latency}
show the training and prediction latency times of each implementation
versus the numbers of input and hidden nodes (all the reported times are averages over 50,000 trials).
To measure practical latency times,
the exploration range of the number of input nodes is set to \{128, 256, 512, 1,024\},
while that of the number of hidden nodes is set to \{16, 32, 64\} on the basis of the hyperparameter settings of {\detector} in Section \ref{sec:eval_anomaly}.
\begin{table}[t!]
    \centering
    \scriptsize
    \caption{Hyperparameter Settings in Section \ref{sec:eval_hard}}
    \label{tb:hard_hp}
    \begin{tabular}{c|c|c|c}\hline\hline
        \begin{tabular}{c} {\core} \\ $\{G_{hidden}, \tilde{N}_1, L\}$ \end{tabular} & \begin{tabular}{c} NN-AE-\{*\} \\ $\{G_{hidden}, G_{out}, \tilde{N}_1, L, O, B\}$ \end{tabular} & \begin{tabular}{c} DNN-AE-\{*\} \\ $\{G_{hidden}, G_{out}, \tilde{N}_1, \tilde{N}_2, \tilde{N}_3, L, O, B\}$ \end{tabular} & \begin{tabular}{c} FPELM-AE-\{*\} \\ $\{G_{hidden}, \tilde{N}_1, L\}$ \end{tabular} \\\hline
        $\{$Identity, $\tilde{N}$, MSE$\}$ & $\{$Relu, Sigmoid, $\tilde{N}$, MSE, Adam, 1$\}$ & $\{$Relu, Sigmoid, $2\tilde{N}$, $\tilde{N}$, $2\tilde{N}$, MSE, Adam, 1$\}$ & $\{$Identity, $\tilde{N}$, MSE$\}$ \\\hline
    \end{tabular}
\end{table}
\begin{figure}[t!]
    \begin{minipage}{0.33\hsize}
        \centering
        \includegraphics[height=35.0mm]{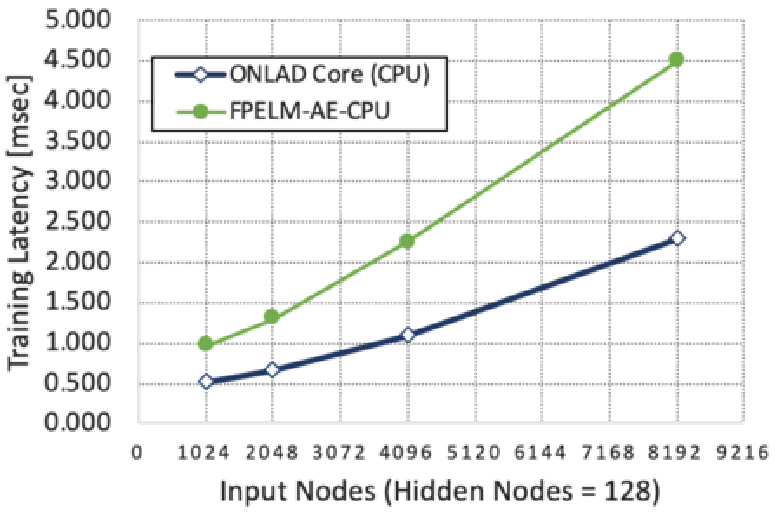}
    \end{minipage}
    \begin{minipage}{0.33\hsize}
        \centering
        \includegraphics[height=35.0mm]{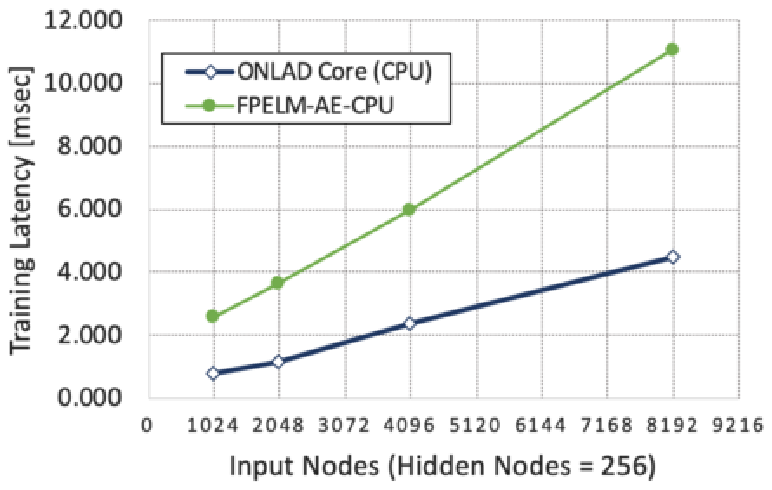}
    \end{minipage}
    \begin{minipage}{0.33\hsize}
        \centering
        \includegraphics[height=35.0mm]{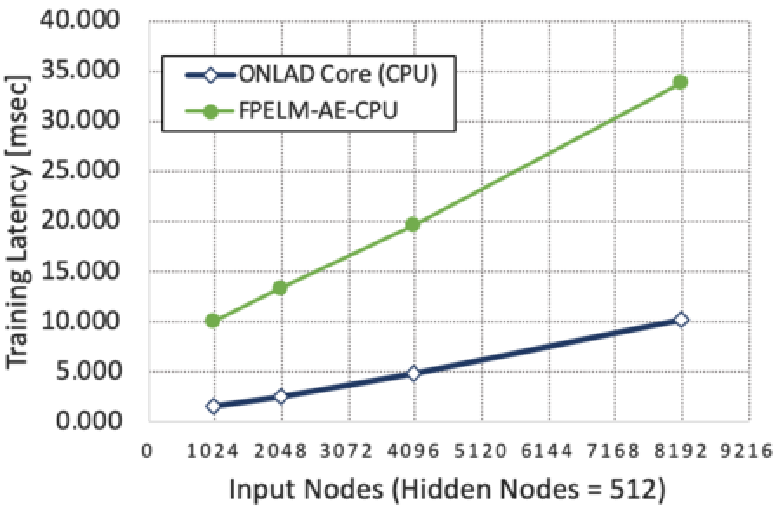}
    \end{minipage}
    \caption{Comparison of Training Latency of Proposed Forgetting Mechanism and FP-ELM}
    \label{fig:forget_latency}
\end{figure}

As shown in the figures, the latency times of the software implementations remain almost constant as the number of input nodes increases.
This outcome shows that most of their execution times are occupied with software overheads to invoke training and prediction tasks.
The GPU-based implementations especially suffer from high latency times because of the communication cost between a GPU and a CPU in addition to the software overheads.
In contrast, {\core} is free from these overheads.
Consequently, {\core} achieves 1.95x, 2.45x, 2.56x, 3.38x, 4.51x, and 6.58x speedups on average over {\nncpu}, {\dnncpu}, {\fpelmcpu}, {\nngpu}, {\dnngpu}, and {\fpelmgpu} in terms of training latency,
and 2.29x, 2.37x, 2.36x, 4.38x, 4.73x, and 4.57x speedups on average over them in terms of prediction latency.
{\core} can perform fast sequential learning and prediction to follow concept drift approximately in less than one millisecond.

However, please note that {\core} may become slower than the others when there are many input nodes
since the computational cost of Train/Predict Module is proportional to the number of input nodes, as shown in Equations \ref{eq:train_iter} and \ref{eq:predict_iter}.
Hence, {\core} has difficulty achieving speedups beyond 1.0x over the software implementations when there are thousands of input nodes.
\begin{figure}[t]
    \centering
    \includegraphics[width=65.0mm]{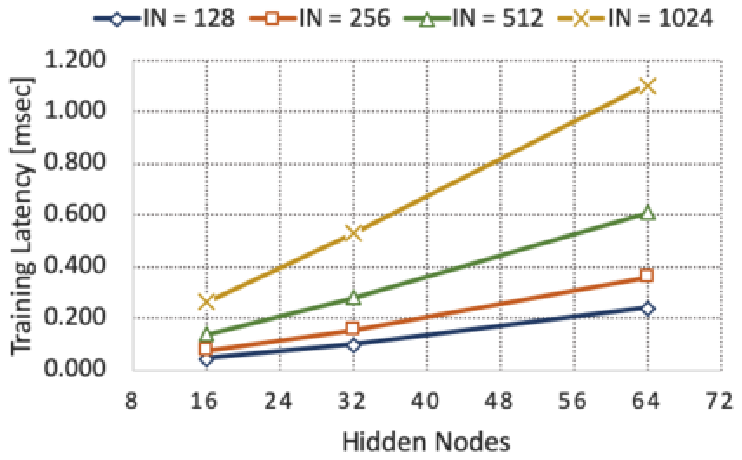}
    \caption{Relationship between Training Latency of {\core} and Hidden Nodes}
    \label{fig:train_latency_hv}
\end{figure}

Moreover, the computational cost of Train Module is proportional to the square of the number of hidden nodes, too.
However, contrary to expectations, Figure \ref{fig:train_latency_hv} shows that the latency times are almost proportional to the number of hidden nodes.
This is because $(3n + 1)\tilde{N} \gg 4\tilde{N}^2$ holds as long as $n \gg \tilde{N}$ in Equation \ref{eq:train_iter}.
In other words, the computational cost of Train Module stays almost proportional to the number of hidden nodes
as long as the number of input nodes is much greater than that of hidden nodes.
The practicality of this condition is empirically demonstrated;
the best hyperparameter settings of {\core} satisfy $n \gg \tilde{N}$ as shown in Tables \ref{tb:eval_offline_hp} and \ref{tb:eval_online_hp}.
Hence, in practical situations, the computational cost of {\core} does not excessively increase
even when the number of hidden nodes is increased.

\subsubsection{Computational Cost of Proposed Forgetting Mechanism}
Here, the proposed forgetting mechanism of {\core} and the baseline algorithm (i.e., FP-ELM) are
compared in terms of computational cost.
Since the forgetting operation of {\core} or FP-ELM is unified
into the training algorithm, we use training latency times to compare them.
Also, to make a fair comparison of their computational costs,
we compare a CPU implementation of {\core} and {\fpelmcpu},
both of which are implemented with the same library (i.e., Tensorflow).

Figure \ref{fig:forget_latency} shows the experimental results,
where the exploration ranges of input and hidden nodes are set to 8x larger
than those of Figures \ref{fig:train_latency} and \ref{fig:predict_latency},
in order to increase the ratio of computation time of the models
and make a clear comparison of their computational costs.
Consequently, our forgetting mechanism is faster than
{\fpelmcpu} by 3.21x on average.
The computational cost of our forgetting mechanism is $O(\tilde{N}^2)$ as shown in Equation \ref{eq:train_iter},
however, that of FP-ELM is $O(\tilde{N}^3)$ since the matrix size of the matrix inversion of FP-ELM is $\tilde{N} \times \tilde{N}$;
the gap of their computation times gradually widens as the number of hidden nodes increases.

\subsection{FPGA Resource Utilization}\label{ssec:eval_hard_resource}
This section evaluates FPGA resource utilization of {\core} by varying the numbers of input and hidden nodes.
The exploration range of the number of input nodes is chosen to be \{128, 256, 512, 1,024\},
and that of the number of hidden nodes is to \{16, 32, 64\} on the basis of the results in the previous section.
For ease of analysis, we use pre-synthesis resource utilization reports produced by Vivado HLS as experimental results.

\begin{figure}[t]
    \begin{minipage}{0.5\textwidth}
        \centering
        \scriptsize
        \makeatletter
        \def\@captype{table}
        \caption{FPGA Resource Utilization of {\core} (Pre-synthesis Results)}
        \makeatother
        \begin{tabular}{c|c|c|c|c} \hline\hline
            \multicolumn{5}{c}{Hidden Nodes = 16} \\ \hline
            Input Nodes & BRAM [\%] & DSP [\%] & FF [\%] & LUT [\%] \\ \hline
            128 & 10.0 & 40.0 & 16.0 & 29.9 \\
            256 & 12.9 & 40.0 & 16.1 & 30.0 \\
            512 & 18.6 & 40.0 & 16.1 & 30.0 \\
            1,024 & 30.0 & 40.0 & 16.1 & 30.0 \\ \hline
            \multicolumn{5}{c}{Hidden Nodes = 32} \\ \hline
            Input Nodes & BRAM [\%] & DSP [\%] & FF [\%] & LUT [\%] \\ \hline
            128 & 13.6 & 40.0 & 16.0 & 29.9 \\
            256 & 19.3 & 40.0 & 16.0 & 30.0 \\
            512 & 30.7 & 40.0 & 16.0 & 30.0 \\
            1,024 & 53.6 & 40.0 & 16.0 & 30.0 \\ \hline
            \multicolumn{5}{c}{Hidden Nodes = 64} \\ \hline
            Input Nodes & BRAM [\%] & DSP [\%] & FF [\%] & LUT [\%] \\ \hline
            128 & 24.3 & 40.0 & 15.9 & 30.0 \\
            256 & 35.7 & 40.0 & 15.9 & 30.0 \\
            512 & 58.6 & 40.0 & 16.0 & 30.0 \\
            1,024 & \textbf{104.2} & 40.0 & 16.0 & 30.1 \\ \hline
        \end{tabular}
        \label{tb:resource}
    \end{minipage}
    \begin{minipage}{0.5\textwidth}
        \centering
        \includegraphics[width=70.0mm]{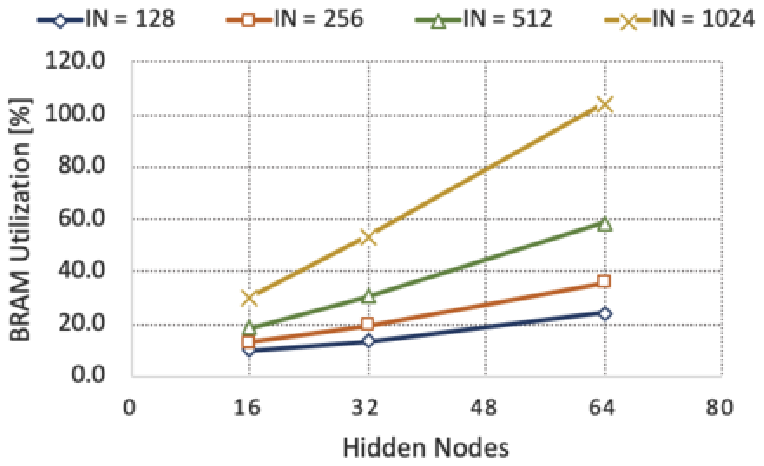}
        \caption{Relationship between BRAM Utilization and Hidden Nodes}
        \label{fig:area_hv}
    \end{minipage}
\end{figure}

Table \ref{tb:resource} shows the experimental results.
The DSP utilization remains almost constant even as the numbers of input and hidden nodes increase.
This is a reasonable outcome since the DSP slices are consumed only for Train Module and Predict Module,
and both of them are designed to use a specific number of arithmetic units regardless of the number of input and hidden nodes,
as mentioned in Sections \ref{sssec:train_module} and \ref{sssec:predict_module}.

However, {\core} consumes more BRAM instances as the model size increases.
$S_{onlad}$ below denotes the total number of matrix elements of the entire {\core}.
\begin{equation}\label{eq:onlad_space}
    \begin{split}
        S_{onlad} &= S_{parameter} + S_{input} + S_{train} + S_{predict} \\
        &= 5\tilde{N}^2 + (5n + 4)\tilde{N} + 2n + 1
    \end{split}
\end{equation}
Equation \ref{eq:onlad_space} shows that the utilization of the BRAM instances of {\core} is linearly increased
as the number of input nodes $n$ increases.
The experimental results shown in Table \ref{tb:resource} are consistent with Equation \ref{eq:onlad_space};
the BRAM utilization is proportional to the number of input nodes.

Equation \ref{eq:onlad_space} also shows that the utilization of the BRAM instances is
proportional to the square of the number of hidden nodes $\tilde{N}^2$, too.
However, the BRAM utilization ratios of {\core} are almost proportional to the number of hidden nodes,
as shown in Figure \ref{fig:area_hv}.
The same logic as in the previous section can explain this outcome;
$(5n + 4)\tilde{N} \gg 5\tilde{N}^2$ holds as long as $n \gg \tilde{N}$ in Equation \ref{eq:onlad_space}.
The practicality of the condition $n \gg \tilde{N}$ is also as described in the previous section.
Hence, in practical situations, the BRAM utilization of {\core} is suppressed and does not excessively increase
even if the number of hidden nodes increases.
Consequently, except for the largest setting $(n, \tilde{N}) = (1,024, 64)$,
all the utilization rates of {\core} are under the limit.

\subsection{Power Consumption}\label{ssec:seval_hard_power}
This section evaluates the runtime power consumption of
our board-level implementation in comparison with the other software implementations.
We use an ordinary watt-hour meter to measure the power consumption of PYNQ-Z1 board.
For the software implementations, s-tui and nvidia-smi are used.
s-tui \cite{s-tui} is an open-source CPU monitoring tool; we use it to
measure the power consumption of the CPU (i.e., Intel Core i7 6700 3.4GHz) equipped in the experimental machine.
On the other hand, nvidia-smi is a GPU monitoring utility provided by Nvidia.
We use it to measure the power consumption of the GPU (i.e., Nvidia GTX 1070 8GB) equipped in the experimental machine.

The input and hidden nodes of all the implementations are commonly set to 512 and 64.
This setting has been confirmed to consume the largest amount of resources in Table \ref{tb:resource}.
The resource utilization report of {\core} is shown in Table \ref{tb:resource_onboard}.

\begin{figure}[t]
    \centering
    \includegraphics[width=65.0mm]{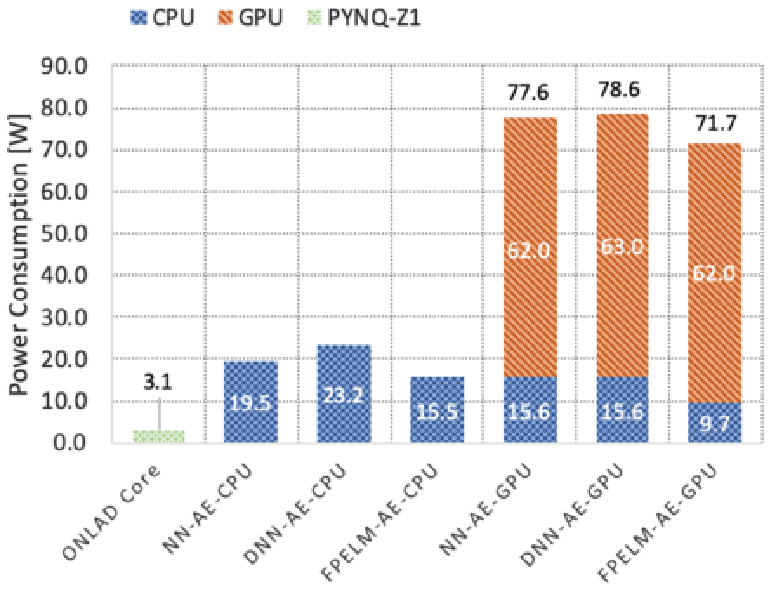}
    \caption{Comparison of Power Consumption}
    \label{fig:power}
\end{figure}

Figure \ref{fig:power} shows the power consumption of each implementation when training computations are continuously executed.
As shown in the figure, our implementation consumes 3.1 W, 5.0x$\sim$25.4x lower than the others.
The reported power consumption of our implementation includes not only that of {\core}, but also that of other components such as a dual-core ARM CPU.
Hence, the power consumption of {\core} itself is even lower than 3.1 W.
\section{Related Work}\label{sec:related}

\subsection{On-device Learning}
Data play an important role in machine learning, although sometimes they can be privacy-sensitive.
Here, on-device prediction/learning is a way to ensure data privacy because it does not require user data transfers with external server machines.
Ravi \textit{et al}. proposed ProjectionNet \cite{projection_net} to make existing BP-NN-based models smaller and reduce
the memory they take up on user devices without significantly degrading accuracy.
This is done by leveraging an LSH (Locality Sensitive Hashing) based projection method and a distillation training framework.
Kone\v{c}n\'{y} \textit{et al}. proposed a federated learning framework \cite{federated_optim}, which utilizes user devices as computational nodes to train a global model.
In this framework, user devices are supposed to perform training only with their local data; then, the updated weights are aggregated into the global model.
Zhu \textit{et al}. and Park \textit{et al}. studied federated learning approaches with edge devices on wireless sensor networks \cite{edge_learning_zhu}\cite{edge_learning_park}.
They explored their essential building blocks and pointed out underlying challenges.
Our approach shares a common idea that edge devices themselves perform training,
although the aim is not to create a global model; our work tries to create a locally personalized model for the target edge device.

\subsection{Anomaly Detection with OS-ELM}
Since sequential learning approaches are capable of learning input data online,
they have been utilized for anomaly detection where
real-time adaptation and prediction are often required.
OS-ELM is no exception; several studies have been reported on anomaly detection using OS-ELM.
Nizar \textit{et al}. proposed an OS-ELM-based irregular behavior detection system of electricity customers to
prevent non-technical losses such as power theft and illegal connections \cite{nizar}.
They compared their system with SVM based ones and showed its superiority.
Singh \textit{et al}. proposed an OS-ELM-based network traffic IDS (Intrusion Detection System).
They showed that the system can perform training on a huge amount of traffic data even with limited memory space \cite{singh_ids}.
Bosman \textit{et al}. proposed a decentralized anomaly detection system for wireless sensor networks \cite{bosman}.
On the other hand, we utilize OS-ELM for semi-supervised anomaly detection in conjunction with an autoencoder.
As far as we know, we propose the combination as the first work.

\subsection{OS-ELM Variants with Forgetting Mechanisms}
Over the past several years, several OS-ELM variants with forgetting mechanisms have been proposed.
Zhao \textit{et al.} were the first to study a forgetting mechanism for OS-ELM, called FOS-ELM \cite{fos_elm}.
FOS-ELM takes a sliding-window approach, where the latest $s$ training chunks are taken into account ($s$ is a fixed parameter of window size).
On the other hand, $\lambda_{DFF}$OS-ELM \cite{lambda_dff_os_elm} and FP-ELM \cite{fp_elm} introduce variable forgetting factors to forget old training chunks gradually.
They adaptively update the forgetting factors according to the information in arriving input data or output error values.
Our approach is based on FP-ELM, though it is modified to provide the forgetting mechanism with a tiny additional computational cost to the original algorithm of OS-ELM.
\begin{table}[t]
    \centering
    \footnotesize
    \caption{FPGA Resource Utilization of {\core} (Post-synthesis Result)}
    \label{tb:resource_onboard}
    \begin{tabular}{c|c|c|c} \hline\hline
        BRAM [\%]& DSP [\%]& FF [\%]& LUT [\%] \\\hline
        55.4 & 32.7 & 11.6 & 25.8 \\\hline
    \end{tabular}
\end{table}

\subsection{Hardware Implementations of OS-ELM}
Several papers on hardware implementations of ELM \cite{elm_hard1}\cite{elm_hard2}\cite{elm_hard3}\cite{elm_hard4} have been reported since 2012.
However, implementations of OS-ELM have just started to be reported.
Tsukada \textit{et al}. provided a theoretical analysis for hardware implementations of OS-ELM to significantly reduce the computational cost \cite{os_elm_fpga_tsukada}.
Villora \textit{et al}. and Safaei \textit{et al}. proposed fast and efficient FPGA-based implementations of OS-ELM for embedded systems \cite{os_elm_fpga_villora}\cite{os_elm_fpga_safaei}.
In this paper, we propose an IP core that implements the proposed OS-ELM-based semi-supervised anomaly detection approach.
This IP core can be implemented on edge devices of limited resources and works at low power consumption.

\subsection{Neural Network Based Hardware Implementations for Anomaly Detection}
In this section, we compare several NN-based anomaly detection hardware implementations in Table \ref{tb:nnhw}.
Akin \textit{et al.} proposed an FPGA based condition monitoring system, whose prediction time is less than 2 msec, for induction motors \cite{nn_anoma_fpga_akin}.
The proposed system employs a supervised anomaly detection approach using a 3-layer binary-classification model;
it requires both anomaly data and normal data for training.
Wess \textit{et al.} proposed an electrocardiogram anomaly detection approach based on FPGA \cite{nn_anoma_fpga_wess}.
The proposed system consists of (1) feature extraction, (2) dimensional reduction, and (3) classification,
in which (3) is implemented as a dedicated circuit on FPGA.
They reported that the prediction latency is approximately less than 100 cycles,
although their approach is also based on a classification model as well as \cite{nn_anoma_fpga_akin}.
In contrast to the above implementations, {\core} adopts a semi-supervised approach, where only normal data are required for training.

Moss \textit{et al.} proposed an FPGA based anomaly detector for radio frequency signals \cite{nn_anoma_fpga_moss}.
The proposed IP core realizes semi-supervised anomaly detection using a BP-NN based autoencoder, which is a similar approach to our work.
Also, its prediction latency is as fast as 105 nsec.
However, the model size (i.e., weight parameters) is 100x smaller than {\core},
and the FPGA platform is much larger than ours.
Besides, their IP core does not support training computations.
Alrawashdeh \textit{et al.} proposed a DBN (Deep Belief Network) based IP core that supports training as with {\core} for anomaly detection \cite{nn_anoma_fpga_alrawashdeh}.
They proposed a cost-efficient training model for the contrastive divergence algorithm of DBN
and reported that the performance of the IP core achieves 37 Gops/W.
However, the model adopts a classification based approach as with \cite{nn_anoma_fpga_akin} and \cite{nn_anoma_fpga_wess}.
On the other hand, {\core} supports training,
and at the same time it adopts a semi-supervised anomaly detection approach,
which makes it more applicable to a wide range of real-world applications.
\begin{table}[t]
    \centering
    \scriptsize
    \caption{Comparison of NN-based Hardware Implementations for Anomaly Detection}
    \label{tb:nnhw}
    \begin{tabular}{c|c|c|c|c|c}\hline\hline
    & Akin \textit{et al.} \cite{nn_anoma_fpga_akin} & Wess \textit{et al.} \cite{nn_anoma_fpga_wess} & Moss \textit{et al.} \cite{nn_anoma_fpga_moss} & Alrawashdeh \cite{nn_anoma_fpga_alrawashdeh} & {\core}\\\hline
    Approach & \begin{tabular}{c} supervised \\ (classification) \end{tabular} & \begin{tabular}{c} supervised \\ (classification) \end{tabular} & \begin{tabular}{c} semi-supervised \\ (autoencoder) \end{tabular} & \begin{tabular}{c} supervised \\ (classification) \end{tabular} & \begin{tabular}{c} semi-supervised \\ (autoencoder) \end{tabular} \\
    NN Model & BP-NN & BP-NN & BP-NN & DBN & OS-ELM \\
    Layers & 3 & 3 & 5 & 4 & 3 \\
    Weight Parameters & \begin{tabular}{c} 12 \end{tabular} & \begin{tabular}{c} $\sim$84 \end{tabular} & \begin{tabular}{c} 1,280 \end{tabular}& N/A & \begin{tabular}{c} $\sim$131,072 \end{tabular}\\ 
    Platform & Altera Cyclone iii Devkit & Avnet Zedboard & Ettus USRP X310 & Xilinx ZC706 & Digilent PYNQ-Z1 \\
    Tools & Quartus ii (VHDL) & Vivado (HLS) & Vivado (HLS) & Vivado (Verilog) & Vivado (HLS) \\
    \begin{tabular}{c} Training Supported ? \end{tabular} & No & No & No & Yes & Yes \\
    Frequency & 50 MHz & N/A & 200 MHz & N/A & 100 MHz \\
    Prediction Latency & $\sim$2 msec & $\sim$100 cycles & 105 nsec & 8$\mu$sec& $\sim$1 msec\\
    Training Latency & N/A & N/A & N/A & N/A & $\sim$1 msec\\
    Power Consumption & N/A & N/A & N/A & N/A & 3.1 W \\
    Power Efficiency & N/A & N/A & N/A & 37 Gops/W & N/A \\\hline
    \end{tabular}
\end{table}

\subsection{Design Tools for Hardware Implementation of Neural Networks}
The PYNQ-Z1 board used in this work provides the PYNQ library \cite{pynq_library}
which allows the developers to design CPU-FPGA co-architecture with Python codes,
although it is not specialized for implementing neural networks.
fpgaConvNet \cite{fpgaconvnet} is an automated design framework
for CNN (Convolutional Neural Network) based classification models on FPGA platforms.
This framework adopts a synchronous dataflow model where
the design space of performance and cost is explored, while taking into account platform-specific constraints.
DnnWeaver \cite{dnnweaver} is a design tool that generates synthesizable DNN accelerators
from high-level configurations in Caffe.
The DnnWeaver compiler tiles, schedules, and batches DNN operations to maximize data reuse and utilize target FPGA's memory.
Zhao \textit{et al.} proposed a high-level design framework for BNNs (Binarized Neural Networks) \cite{progbnn}.
Since the main arithmetics of BNNs are simple bitwise logic operations
instead of costly floating-point operations,
the computational cost and FPGA resources required to implement the accelerator can be significantly reduced
compared with conventional CNNs.
GUINNESS \cite{guiness} is a GUI based design tool for implementing BNNs on Xilinx SoC platforms.
In this tool, the designers do not need to write any RTL codes or scripts,
which enables software designers to develop prototypes of BNN-based accelerators without knowledge of hardware.
\section{Conclusions}\label{sec:conc}

\subsection{Summary}\label{ssec:summary}
In this work, we proposed {\detector}
which realizes fast sequential learning semi-supervised anomaly detection
by constructing an autoencoder with OS-ELM.
We showed that the computational cost of OS-ELM is significantly reduced when the batch size is fixed to 1,
which contributes to speedup of {\detector}.
Also, we proposed a computationally lightweight forgetting mechanism
for OS-ELM, based on FP-ELM.
It enables {\detector} to follow concept drift at a low computational cost.
In addition, we proposed {\core}
in order to realize on-device execution of {\detector} on resource-limited edge devices at low power consumption.
Since {\core} does not need to offload training computations to external remote server machines,
it enables standalone execution where no data transfers to server machines are required.

Experimental results using public datasets showed that {\detector} has comparable
generalization capability to that of BP-NN-based models in the context of anomaly detection.
We also confirmed that {\detector} has favorable anomaly
detection capability especially in an environment that simulates concept drift.

Evaluations of {\core} confirmed that it can perform training and prediction computations faster than software implementations of BP-NNs and FP-ELM by 1.95x$\sim$6.58x and 2.29x$\sim$4.73x on average.
They also comfirmed that the proposed forgetting mechanism is faster than FP-ELM by 3.21x on average.
In addition, our evaluations showed that {\core} can be implemented on PYNQ-Z1 board
in practical model sizes.
We demonstrated that the runtime power consumption of PYNQ-Z1 board that
implements {\core} is 5.0x$\sim$25.4x lower in comparison with the other software implementations when
training computations are continuously executed.

\subsection{Future Directions}\label{ssec:future}
BP-NNs are known to achieve higher generalization performance to some extent by stacking more layers.
Although the original OS-ELM algorithm is limited to have only one hidden layer,
ML-OSELM (Multi-Layer Online Sequential Extreme Learning Machine) \cite{ml_os_elm} proposed by Mirza \textit{et al.}
provides a multi-layer framework for OS-ELM.
According to \cite{ml_os_elm}, ML-OSELM outperforms OS-ELM on well-known open classification datasets
by 0.15$\sim$2.58\% in terms of test accuracy.
Thus, anomaly detection capability of {\detector} can be further improved by replacing OS-ELM in {\detector} with ML-OSELM.
We plan to work with the multi-layer version of {\detector} and {\core}.

In real world, there are some systems that have multiple action modes such as air conditioners, robot arms, and gas turbines.
In the context of anomaly detection, such systems are often formulated as mixture models which consist of multiple sub-distributions of normal data.
Recently, a mixture model framework that utilizes multiple OS-ELM instances was proposed in \cite{itoh}.
We plan to apply this framework to {\detector} and {\core}.

\bibliographystyle{unsrt}

\end{document}